\documentclass[10pt,twocolumn,letterpaper]{article}

\usepackage{cvpr}              %

\usepackage{graphicx}
\usepackage{amsmath}
\usepackage{amssymb}
\usepackage{booktabs}
\usepackage{algorithm}
\usepackage{algorithmic}
\usepackage{appendix}
\usepackage{soul}

\usepackage[pagebackref,breaklinks,colorlinks]{hyperref}

\newcommand{\real}{\mathbb{R}}
\newcommand{\figref}[1]{Fig.~\ref{#1}}

\newcommand{\secref}[1]{Sec.~\ref{#1}}
\newcommand{\eqnref}[1]{Eqn.~\ref{#1}} 
\newcommand{\algref}[1]{Alg.~\ref{#1}}

\usepackage[capitalize]{cleveref}
\crefname{section}{Sec.}{Secs.}
\Crefname{section}{Section}{Sections}
\Crefname{table}{Table}{Tables}
\crefname{table}{Tab.}{Tabs.}

\def\cvprPaperID{4302} %
\def\confName{CVPR}
\def\confYear{2022}

\begin{document}

\title{Stereoscopic Universal Perturbations across Different Architectures and Datasets}

\renewcommand\footnotemark{}
\renewcommand\footnoterule{}

\author{Zachary Berger$^\dagger$ \\
UCLA Vision Lab\\
{\tt\small zackeberger@g.ucla.edu}
\and
Parth Agrawal$^\dagger$ \\
UCLA Vision Lab\\
{\tt\small parthagrawal24@g.ucla.edu}
\and
Tian Yu Liu \\
UCLA Vision Lab\\
{\tt\small tianyu139@g.ucla.edu}
\and
Stefano Soatto \\
UCLA Vision Lab\\
{\tt\small soatto@cs.ucla.edu}
\and
Alex Wong \\
UCLA Vision Lab\\
{\tt\small alexw@cs.ucla.edu}
\thanks{$^\dagger$ denotes authors with equal contributions.}
\thanks{Code: {\color{blue} github.com/alexklwong/stereoscopic-universal-perturbations}}
}

\maketitle

\begin{abstract}
   We study the effect of adversarial perturbations of images on deep stereo matching networks for the disparity estimation task. We present a method to craft a single set of perturbations that, when added to any stereo image pair in a dataset, can fool a stereo network to significantly alter the perceived scene geometry. Our perturbation images are ``universal'' in that they not only corrupt estimates of the network on the dataset they are optimized for, but also generalize to different architectures trained on different datasets. We evaluate our approach on multiple benchmark datasets where our perturbations can increase the D1-error (akin to fooling rate) of state-of-the-art stereo networks from 1\% to as much as 87\%. We investigate the effect of perturbations on the estimated scene geometry and identify object classes that are most vulnerable. Our analysis on the activations of registered points between left and right images led us to find architectural components that can increase robustness against adversaries. By simply designing networks with such components, one can reduce the effect of adversaries by up to 60.5\%, which rivals the robustness of networks fine-tuned with costly adversarial data augmentation. Our design principle also improves their robustness against common image corruptions by an average of 70\%.
\end{abstract}

\begin{figure*}[ht]
    \centering
    \includegraphics[width=1.0\textwidth]{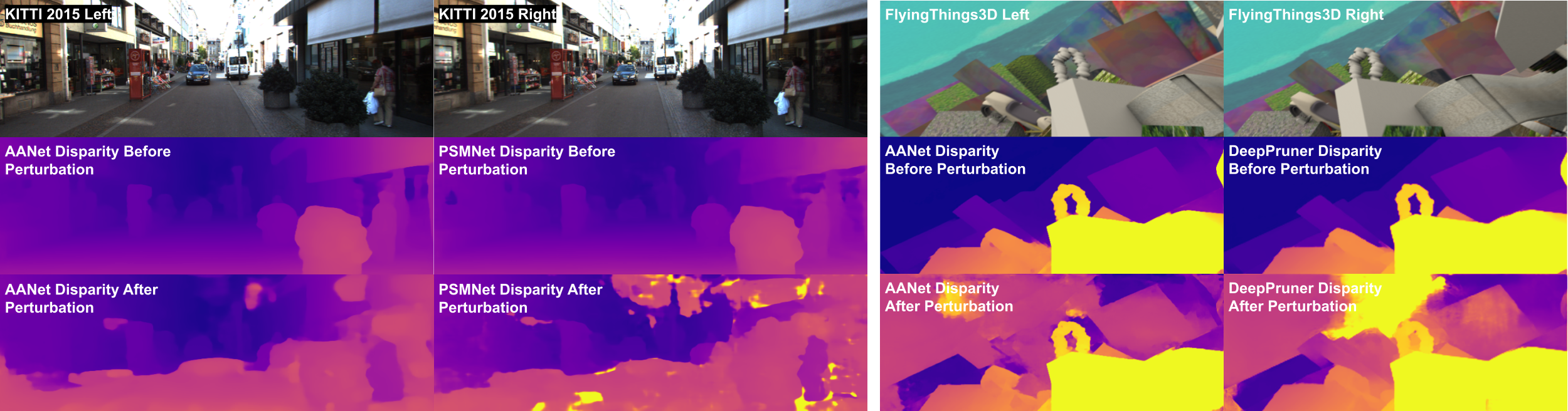}
    \caption{\textit{Universality across models and datasets}. We optimized a single pair of perturbation images for AANet on the KITTI dataset. When added to a stereo pair from KITTI 2015, it corrupts the disparity estimates of AANet and PSMNet. The same perturbations can be added to stereo pairs in Flyingthings3D to fool AANet and DeepPruner.}
\label{fig:teaser}
\vspace{-1em}
\end{figure*}

\section{Introduction}
\label{sec:introduction}
Deep neural networks are vulnerable to adversarial perturbations, where small changes in the input image(s) can cause large inference errors, for instance in the label of objects portrayed within. Even when the images contain sufficient information for inference, for instance in stereo where the disparity between two calibrated images is used to infer the depth of the scene, adversarial perturbations have been shown to alter the depth map \cite{wong2020stereopagnosia}. Such perturbations are ordinarily specific to each individual input, and depend on the particular deep network architecture and the particular dataset on which it is trained. 

For classification, \cite{moosavi2017universal} showed that a single perturbation can be crafted to disrupt the inference for all images in a dataset with high probability. These are called ``universal'' adversarial perturbations, even though they are universal to each image in a particular dataset, and usually do not extend to different datasets. In this paper, we show the existence of stereoscopic universal perturbations (SUPs). SUPs are perturbations that can disrupt the depth or disparity estimate of different stereo networks, with different architectures, trained on different datasets, and operating on different images and domains.

Adversarial perturbations arose mainly as a means to study the topology and geometry of the decision boundary of deep networks. Since individual perturbations had to be crafted for each image, security concerns were far fetched. Universal adversarial perturbations, however, revealed vulnerabilities that could be shared among different images. Still, crafting them required knowledge of the architecture and availability of the training set. In contrast, the existence of universal adversarial perturbations for stereo and other spatial inference tasks, common in robotics and autonomy, suggests that such perturbations could present a concern, especially if they can be applied to different images, processed by different neural network models that are trained on different datasets. To the best of our knowledge, we are the first to show, for stereo, that universal perturbations can be applied effectively \textit{even without} knowledge of the trained model, and generalize across domains and datasets. Such perturbations can be optimized on an off-the-shelf model and realized as a filter to be placed on top of a camera lens.

Our main methodological innovation is to design SUPs so that they are approximately space equivariant. We build the perturbation out of a single tile, repeated periodically. Although the tile is not constrained to have periodic boundary conditions, we notice that the model learns perturbations where boundary artifacts are not obvious, partly because the perturbation itself is designed to be small enough to be quasi-imperceptible. Our design naturally regularizes the tile with data, allowing it to generalize to different image pairs, processed with different architectures trained with different datasets -- increasing error from 1\% to as much as 87\% when added to network inputs.

In our experiments, we observe that the errors in disparity induced by SUPs are more pronounced on certain classes of objects. We conjecture that this is due to said classes exhibiting more homogeneous regions, which are more prone to errors in disparity due to ambiguity. We also found that there is a systematic bias towards closer distance (larger disparity) after perturbations. To study the effect of SUPs on stereo networks, we investigate the activations of left and right feature maps before and after adding perturbations. We validate empirically that the embedding function amplifies the adversarial signal: The embedding of perturbed registered features between the images grows more uncorrelated throughout a forward pass than the embedding of the original or ``clean'' registered features, which ``fools'' a stereo network into estimating incorrect correspondences.

Moreover, we use SUPs to improve robustness in stereo networks. We study the effect that different architectural elements (deformable convolutions, and explicit matching modules) have on mitigating perturbations. We observe that by simply designing networks with these elements (and following standard training protocols), one can reduce the effect of adversaries to a similar degree as fine-tuning a model (that lacks such elements) with adversarial data augmentation. While robustness is increased with fine-tuning, it come at a significant cost in time and compute. In contrast, the proposed architectural design choices can mitigate attacks (60.5\% error reduction), and only require a few lines of code; they also improve robustness against common image perturbations i.e. lossy compression, noise, blur by an average of 70\%. Conclusions are valid for three different architectures, across three datasets. While these are chosen to represent the variety in use today, we cannot exclude that there could be tasks, data and models on which our method to craft perturbations is ineffective, and conversely perturbations that are not mitigated by the methods we propose.

\textbf{Our contributions} include: (i) The design of the first stereoscopic universal perturbations (SUPs) that can not only fool the network they are optimized for, but also other networks across multiple datasets. We perform an empirical analysis on how SUPs affect (ii) the estimated scene geometry, (iii) different object classes, and (iv) the features of registered points in a stereo pair. Our results shed light on how SUPs fool stereo networks and led us to uncover (v) architectural designs, i.e.  deformable convolution and explicit feature matching, that mitigate the effect of SUPs to a similar degree as fine-tuning on them. A discussion of potential negative societal impact is available in Supp. Mat.

\vspace{-0.2em}
\section{Related Work}
\vspace{-0.4em}
\label{sec:related_works}
\textbf{Adversarial perturbations.} \cite{szegedy2013intriguing} showed that small additive signals can significantly alter the output of a classification network. \cite{goodfellow2014explaining} introduced the fast gradient sign method (FGSM). \cite{kurakin2016adversarial,madry2017towards,dong2018boosting} extended FGSM to iterative optimization to boost its potency. \cite{moosavi2016deepfool} found the minimal perturbation to alter the predicted class while \cite{peck2017lower} computed the lower bounds on the perturbation magnitudes required to fool a network. \cite{nguyen2015deep} showed that unrecognizable noise can yield high confidence outputs and \cite{ilyas2019adversarial} attributed adversaries to non-robust features. \cite{xie2019improving} improved their transferability across networks with geometric image augmentations. \cite{naseer2019cross} studied their transferability across datasets.

\cite{moosavi2017universal} proposed \textit{universal} adversarial perturbations, where the same perturbation can be added to any image in a dataset to fool a network. \cite{mopuri2017fast} showed that data independent universal perturbations are transferable across different networks and \cite{mopuri2018generalizable} proposed data-free objectives for crafting them. \cite{poursaeed2018generative, mopuri2018nag, hayes2018learning} use generative models to form universal perturbations. \cite{rampini2021universal} proposed universal attacks on graphs, meshes, and point clouds. For those interested, see \cite{chaubey2020universal} for an extensive survey. We also study universal perturbations, but unlike past works focused on single image based problems, we consider the deep stereo matching, where the latent variable (disparity) is constrained by the stereo pair.

Efforts to defend against adversarial attack include adversarial data augmentation during training \cite{kurakin2016adversarial, tramer2017ensemble}, which can be improved with randomization \cite{xie2017mitigating}. \cite{mummadi2019defending, shafahi2020universal} proposed universal adversarial training, 
\cite{buckman2018thermometer,xiao2019enhancing} gradient discretization, and \cite{pang2019mixup,raff2019barrage,xiao2020one} randomization. \cite{guo2017countering,akhtar2018defense, prakash2018deflecting,samangouei2018defense} performed purification, and \cite{liao2018defense} denoising to rectify the image. \cite{xie2020adversarial} used batch normalization to mitigate perturbations. \cite{chen2021robust} used adversarial learning to improve object detection.

Despite many works on adversarial perturbations for classification, few study dense-pixel prediction tasks e.g. segmentation, optical flow, depth estimation. \cite{xie2017adversarial} showed adversarial perturbations for object detection and segmentation. \cite{hendrik2017universal} proposed universal perturbations for segmentation, while \cite{mopuri2018generalizable} studied them for segmentation and single image depth. \cite{wong2020targeted} showed targeted attacks for single image depth while \cite{dijk2019neural} studied them using images augmented with synthetic vehicles. \cite{zolfi2021translucent} examined translucent patch attacks for object detection, and \cite{ranjan2019attacking} visible patch attacks on optical flow. \cite{schrodi2021causes} proposed defenses against physical attacks for optical flow. \cite{wong2020stereopagnosia} demonstrated adversarial attacks for stereo. Like \cite{wong2020stereopagnosia}, we also consider stereo, but instead, we study universal perturbations and show that the same perturbations generalize across network architectures and datasets. 

\textbf{Deep Stereo Matching.} Early works \cite{zagoruyko2015learning, zbontar2016stereo} replaced hand-crafted features with deep features for more robust matching. Recent works realize the entire stereo pipeline as an inductive bias, from feature extraction to cost matching, into 2D and 3D network architectures. 2D architectures leverage correlation layers for matching. For instance, \cite{mayer2016large} formed a 2D cost volume with correlation over left and right features. \cite{pang2017cascade} extended \cite{mayer2016large} to a cascade residual learning framework. AANet \cite{xu2020aanet} also used correlation, but proposed deformable convolutions \cite{zhu2019deformable} when performing cost aggregation to avoid sampling at discontinuities. 3D architectures use feature concatenation and sparse patch matching. \cite{kendall2017end} concatenated left and right features together to build a 3D cost volume. PSMNet \cite{chang2018pyramid} added spatial pyramid pooling layers and introduced a stacked hourglass architecture. \cite{zhang2019ga} used local and global cost aggregation. DeepPruner \cite{duggal2019deeppruner} proposed differentiable patch matching over deep features to construct their cost volume. 

We demonstrate the existence of universal adversarial perturbations on PSMNet, DeepPruner and AANet. We chose them as architectural exemplars for the stereo matching task. PSMNet represents the modern stereo networks (stacked hourglass, cost volume, 3D convolutions), but uses feature stacking without explicit matching. DeepPruner follows the architecture of PSMNet, but performs explicit matching with PatchMatch \cite{barnes2009patchmatch}. AANet represents the state of the art in 2D architecture and uses correlation. 

\vspace{-0.2em}
\section{Universal Perturbations for Stereo}
\vspace{-0.4em}
\label{sec:method}
\textbf{Formulation.} Let $f_\theta(x_L, x_R) \in \real^{H \times W}$ be a pretrained stereo network that estimates the disparity between the left $x_L$ and right $x_R$ images of a stereo pair and $\mathcal{X}$ be a distribution of stereo pairs that belongs to the set of natural images. Our goal is to craft a single pair of image-agnostic stereoscopic universal perturbation images (SUPs) $v_L, v_R \in [0, 1]^{H \times W \times 3}$ that, when added to $(x_L, x_R)$, corrupts the disparity estimate such that $f_\theta(x_L, x_R)~\neq~f_\theta(\hat{x}_L~,~\hat{x}_R)$~where $\hat{x}_L = x_L+v_L$ and $\hat{x}_R = x_R+v_R$ for $(x_L, x_R) \sim \mathcal{X}$. To ensure that the SUPs are small or quasi-imperceptible, we subject them to the norm constraints $\|v_I\|_{\infty} \leq \epsilon$ for $I \in \{L, R\}$.

We assume a dataset $X := \{(x_L^{(n)}, x_R^{(n)}), y_{gt}^{(n)}\}_{n=1}^{N}$ sampled from $\mathcal{X}$ as a ``training'' set, and access to a stereo network $f_\theta$ and its loss function $\ell(f_\theta(\cdot), y_{gt})$ where $y_{gt}$ denotes the ground truth. We note that, unlike classification or segmentation, it is rare for any large scale stereo dataset to provide ground truth for every sample, so instead we use disparity estimated from ``clean'' stereo pairs, i.e. without any perturbations, as pseudo ground truth, $y^{(n)} = f_\theta(x^{(n)}_L, x^{(n)}_R)$.

\textbf{Algorithm.} To craft universal perturbations subject to the norm constraint of $\|v_I\|_{\infty} \leq \epsilon$, we propose to generate $(v_L, v_R)$ by iterating through $X$ and gradually aggregating small perturbation vectors that are able to fool the stereo network $f_\theta$ into altering its output disparity or the perceived scene geometry for a given image pair $(x_L^{(n)}, x_R^{(n)}) \in X$. At each iteration, we compute the gradient of the loss $\ell$ with respect to each image $x_I$ in the stereo pair for $I \in \{L, R\}$:
\vspace{-0.3em}
\begin{equation}
    g^{(n)}_I = \nabla_{x^{(n)}_{I}} \ell(f_\theta(\hat{x}^{(n)}_L, \hat{x}^{(n)}_R), y^{(n)}).
    \vspace{-0.2em}
\label{eqn:gradient}
\end{equation}
Then, project it onto a smaller (than $\epsilon$) ball with radius $\alpha$ (akin to a learning rate) via the projection operator\footnote{
$\mathbf{p}_{p, \xi}(v) = \arg \min_{v'} \| v - v' \| \ \textrm{subject to} \ \|v'\|_p < \xi$}
and aggregate it to the current perturbations:
\vspace{-0.3em}
\begin{equation}
    v_I = v_I + \mathbf{p}_{\infty, \alpha}(g^{(n)}_I).
    \vspace{-0.2em}
\label{eqn:lp_projection}
\end{equation}
Finally, we project $v_I$ onto the $\epsilon$ radius ball after each iteration to ensure our perturbations meet the upper norm constraint. The procedure is repeated for all stereo pairs in $X$. See \algref{alg:universal_perturbations} for more details.

\setlength{\textfloatsep}{10pt}
\begin{algorithm}[t]
\caption{Computing universal perturbations.}
\begin{algorithmic} 
    \STATE \textbf{Parameters:} Upper norm $\epsilon$, learning rate $\alpha$.
    \STATE \textbf{Inputs:} Dataset $X$, pre-trained stereo network $f_\theta$.
    \STATE \textbf{Outputs:} Perturbations $v_L, v_R$.
    \STATE \textbf{Initialize:} $v_L = \mathbf{0}, v_R = \mathbf{0}$.
    \FOR{each stereo pair $(x^{(n)}_L, x^{(n)}_R) \in X$}
        \STATE Compute $g^{(n)}_L$ and $g^{(n)}_R$ as defined in \eqnref{eqn:gradient}
        \STATE $v_L = \mathbf{p}_{\infty, \epsilon}(v_L + \mathbf{p}_{\infty, \alpha}(g^{(n)}_L))$
        \STATE $v_R = \mathbf{p}_{\infty, \epsilon}(v_R + \mathbf{p}_{\infty, \alpha}(g^{(n)}_R))$
        \ENDFOR
\end{algorithmic} 
\label{alg:universal_perturbations}
\end{algorithm}

\begin{figure*}[ht]
\centering
    \includegraphics[width=1.0\textwidth]{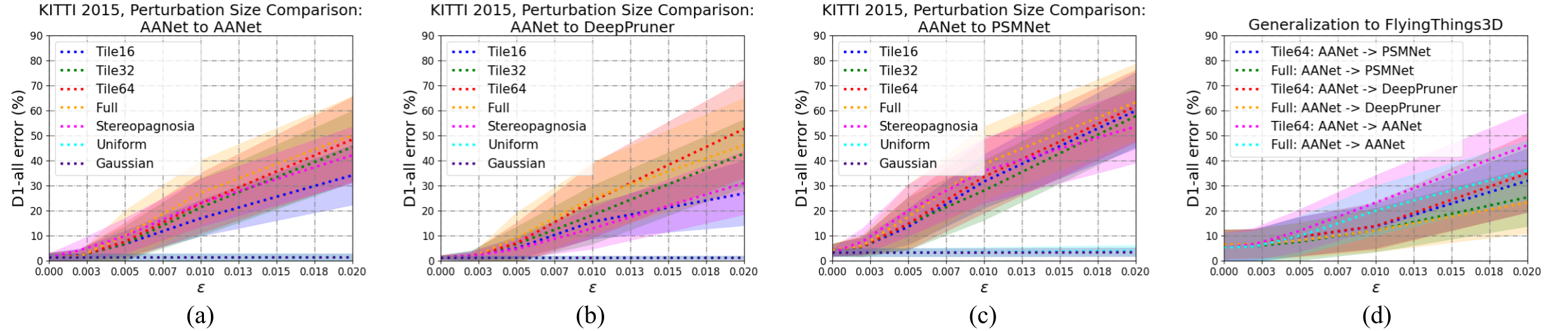}
    \vspace{-1.8em}
    \caption{\textit{The effect of perturbation size}. All methods are robust to naive attacks of uniform and Gaussian noise, as performance is constant across $\epsilon$. Hence, we optimize a pair of perturbations on KITTI for AANet and attack (a) AANet, (b) DeepPruner, and (c) PSMNet. Amongst the perturbation sizes, full and $64 \times 64$ are the most effective at degrading performance on KITTI 2015 validation set. In (d), we show transferability to FlyingThings3D by using the same perturbations optimized on KITTI from (a)-(c) to attack models trained on Scene  Flow. The $64 \times 64$ perturbations generalize the best across datasets.}
\label{fig:vary_perturb_sizes}
\vspace{-1.3em}
\end{figure*}

\textbf{Towards universality across model and data.} We aim to optimize a single pair of perturbations that can alter the perceived geometry of a scene, not just for the network and dataset it is optimized for, but for an array of different unseen network architectures across multiple datasets. To this end, the perturbations must be spatially invariant to generalize across different scene distributions i.e. roads are commonly at the center of the image for outdoor driving scenarios, but a variety of shapes may populate the same region for indoors. Hence, rather than optimizing $(v_L, v_R)$ that span the full $H \times W$ image domain, we reduce $(v_L, v_R)$ to a pair of $h \times w$ patches or tiles subject to $h \mid H$ and $w \mid W$. We note that full size $H \times W$ perturbations are a special case.

To apply $(v_L, v_R)$ to $(x_L, x_R)$ over the entire image space, we evenly repeat or tile the perturbation $v_I$ across $x_I$ with no overlap. Formally, we let $x_I(i, j)$ be the $h \times w$ image region that spans from pixel position $(i, j)$ to $(i + h, j + w)$ for $i \in \{0, \frac{H}{h}, ..., \frac{H(h-1)}{h}\}$ and $j \in \{0, \frac{W}{w}, ..., \frac{W(w-1)}{w}\}$. Thus, the perturbed image region is:
\vspace{-0.2em}
\begin{equation}
    \hat{x}_I(i, j) = x_I(i, j) + v_I \ \ \forall \ \ i, j.
    \vspace{-0.2em}
\label{eqn:tiling}
\end{equation}
We now modify the the gradient computation step in \algref{alg:universal_perturbations} for a given stereo pair $(x^{(n)}_L, x^{(n)}_R)$ by taking the mean over the gradient with respect to the image $g^{(n)}_I$ for all tiles
\vspace{-0.2em}
\begin{equation}
    \bar{g}^{(n)}_I = \frac{h \cdot w}{H \cdot W}\sum_{i, j} g^{(n)}_I(i, j).
    \vspace{-0.2em}
\label{eqn:gradient_tile}
\end{equation}
In doing so, we prevent the perturbations from overfitting to the bias in scene structures induced by the training set e.g. road on bottom of the image and sky on top. We demonstrate in \secref{sec:experiment_results} that this approach yields a single set of universal perturbations that can fool different models across multiple datasets. We note that we can extend our approach to patch attacks by adding the perturbations anywhere on the image, instead of tiling across the image. However, because we constrain our perturbations to be within a small $\epsilon$ ball, unlike \cite{ranjan2019attacking}, a visually imperceptible patch attack is limited in its effect on fooling the network.

\begin{figure*}[t]
\centering
\includegraphics[width=1.0\textwidth]{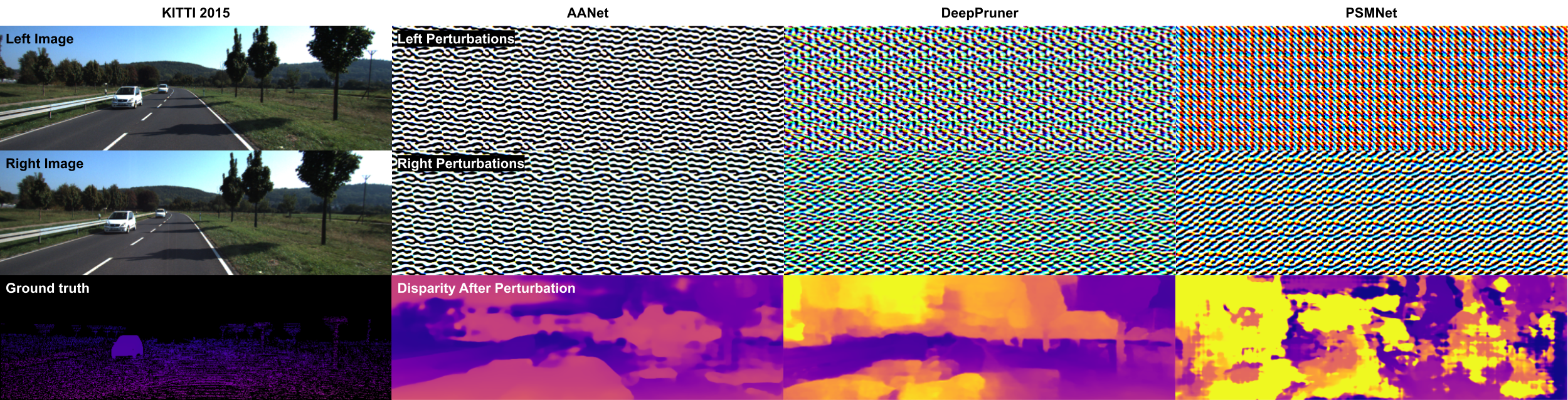}
\vspace{-1.5em}
\caption{\textit{Attacking stereo networks}. We visualize $64 \times 64$ perturbations (tiled across the image domain) optimized for AANet, DeepPruner, and PSMNet on the KITTI dataset. When added to the inputs of the network for which they were optimized, the perturbations can corrupt the estimated disparities. Note: the damage is concentrated on textureless regions e.g. sky, road.}
\label{fig:visualize_perturbations}
\vspace{-1.4em}
\end{figure*}

\vspace{-0.2em}
\section{Experiments and Results}
\vspace{-0.4em}
\label{sec:experiment_results}
We optimized our SUPs on the KITTI raw dataset  \cite{geiger2012we} and evaluated them on KITTI 2012, KITTI 2015 \cite{menze2015object} for stereo models \cite{chang2018pyramid,duggal2019deeppruner,xu2020aanet}. We also show that the same SUPs generalize to FlyingThings3D \cite{mayer2016large} to disrupt models trained on Scene Flow \cite{mayer2016large}. Please see Supp. Mat. for details on datasets, hyper-parameters and implementation.

\begin{figure}[ht]
\centering
\includegraphics[width=1.0\columnwidth]{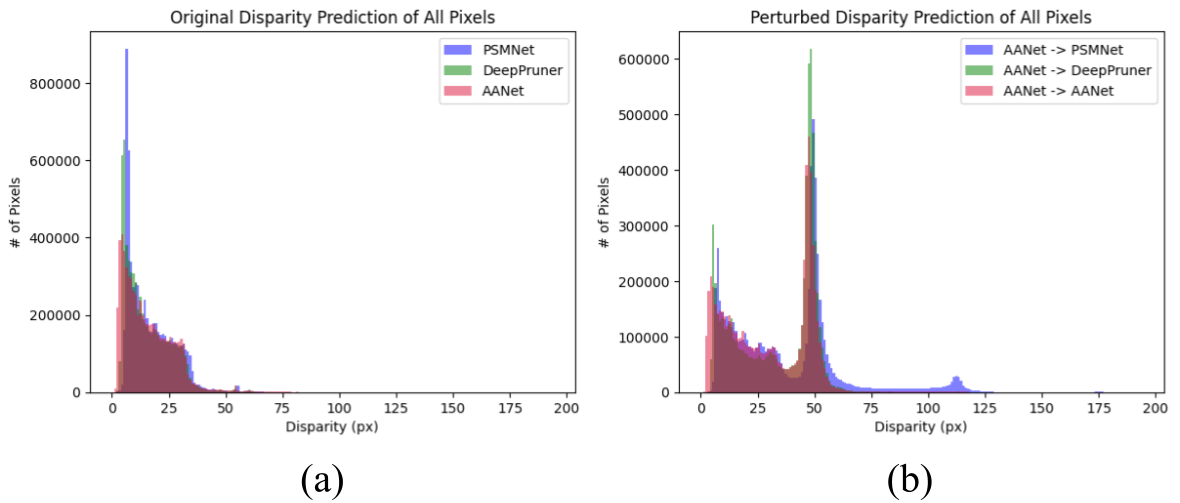}
\vspace{-1.8em}
\caption{\textit{Distribution of disparities before and after adding perturbations}. (a) Before adding perturbations, most of the scene is estimated to be $\approx$2 disparities. (b) The perturbations fool the networks into predicting larger ($\approx$50) disparities.}
\vspace{-0.5em}
\label{fig:aanet_disparity_distribution}
\end{figure}

\textbf{On the effect of perturbation size.} We optimize SUPs on AANet using square tiles of $16, 32$, and $64$, and the full image size of $256 \times 640$. We report results in \figref{fig:vary_perturb_sizes}, which shows the performance of each network on KITTI 2015 when attacked by these perturbations. We compare our results against \cite{wong2020stereopagnosia} which uses image-specific perturbations generated with FGSM. We additionally consider two naive attacks that perturb the input stereo pair $(x_L, x_R)$ with uniform $\mathcal{U}(-\epsilon, \epsilon)$ and Gaussian $\mathcal{N}(0, (\epsilon / 4)^2)$ noise.

\figref{fig:vary_perturb_sizes}-(a, b, c) show that naive attacks have little effect on stereo networks, as the D1-error is roughly constant for all $\epsilon$. Hence, stereo networks are robust to naive perturbations within $\epsilon$ upper norm, and fooling them is non-trivial. Among all square tiles, $64 \times 64$ causes the largest error for all networks across all $\epsilon$. We note that, although our SUPs are image-agnostic, we are comparable to \cite{wong2020stereopagnosia} on small norms and beat them on larger norms. \figref{fig:visualize_perturbations} shows the $64 \times 64$ tiles optimized on KITTI for AANet, DeepPruner, and PSMNet. When added to a stereo pair from KITTI 2015, the disparity estimated by each network is corrupted.

For FlyingThings3D, we consider the full and $64 \times 64$ SUPs (both trained on KITTI) as they caused the most corruption on KITTI 2015. \figref{fig:vary_perturb_sizes}-(d) shows that $64 \times 64$ generalizes better than full-size SUPs across networks. For $\epsilon = 0.02$, $64 \times 64$ achieves $46.14\%$ error on AANet, $34.87\%$ on DeepPruner and $31.93\%$ on PSMNet, while full achieves $36.09\%$, $23.28\%$, and $25.35\%$, respectively. Thus, our tiling approach can help generalize to different data distributions. As our goal is universality across architectures and datasets, we use $64 \times 64$ perturbations for the rest of our experiments.

\begin{figure*}[t]
\centering
\includegraphics[width=0.75\textwidth]{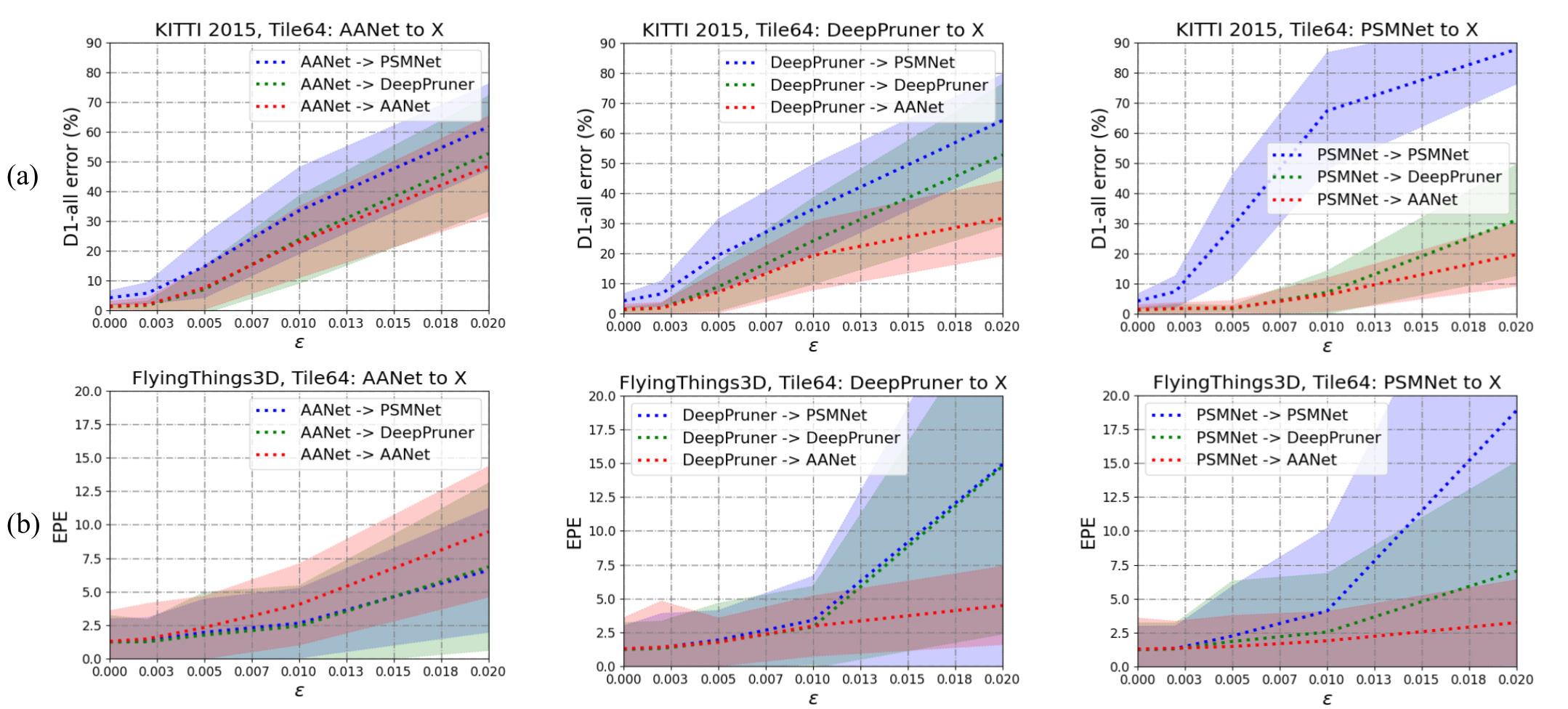}
\vspace{-0.9em}
\caption{\textit{Generalization across architectures and datasets}. Perturbations were optimized for AANet, DeepPruner, and PSMNet on KITTI and added to stereo pairs of KITTI 2015, and FlyingThings3D. Despite being optimized for a specific model on KITTI, they can corrupt models trained on KITTI for KITTI 2015 and those trained on Scene Flow for FlyingThings3D.}
\label{fig:attacking_quantitative_results_main}
\vspace{-1.6em}
\end{figure*}

\textbf{Generalization across architectures and datasets.} We optimized three sets of $64 \times 64$ SUPs on KITTI for AANet, DeepPruner, and PSMNet, respectively. In \figref{fig:attacking_quantitative_results_main}, we attack each network with each set of SUPs on three datasets. We report D1-error for KITTI 2012, and 2015, and EPE for FlyingThings3D (see Supp. Mat. for results for KITTI 2012). For KITTI 2015, \figref{fig:attacking_quantitative_results_main}-(a) shows that, when trained on the network to be attacked, SUPs with $\epsilon = 0.02$ cause error to rise from $1.47\%$ to $48.43\%$ for AANet, $1.28\%$ to $52.74\%$ for DeepPruner, and $4.25\%$ to $87.72\%$ for PSMNet. While SUPs with $\epsilon = 0.002$ have negligible impact, relaxing $\epsilon$ to $0.005$ increases the error of AANet to $7.62\%$, DeepPruner to $8.90\%$, and PSMNet to $28.97\%$. \figref{fig:attacking_quantitative_results_main}-(b) shows that SUPs generalize to other data distributions as well. Adding SUPs optimized on KITTI to FlyingThings3D causes increases in EPE for models trained on Scene Flow -- from 1.30px to 9.47px for AANet, 1.25px to 14.77px for DeepPruner, and 1.27px to 18.88px for PSMNet. 

For all three datasets, our SUPs also generalize across architectures. For example, SUPs with $\epsilon = 0.02$ optimized for AANet on KITTI can be added to stereo pairs in KITTI 2015 to fool DeepPruner (from 1.28\% to 52.66\%), and PSMNet (from 4.25\% to 61.66\%). Similarly, the same SUPs can be added to images in FlyingThings3D to corrupt the outputs of PSMNet (from 1.27 to 6.86px) and DeepPruner (1.25 to 6.60px). Yet, transferability is not symmetric e.g. SUPs optimized for DeepPruner on KITTI only corrupt AANet predictions from 1.30 to 4.49px on Flyingthings3D. \figref{fig:transfer_to_psmnet} demonstrates the transferability qualitatively, showing corruption against PSMNet on FlyingThings3D.

In our experiments, we found AANet to be the most robust and PSMNet the least. We hypothesize that explicit matching plays a role because DeepPruner shares the same architecture as PSMNet, with the exception of a PatchMatch module, but is significantly more robust. Like DeepPruner, AANet also employs matching, but replaces convolutions with deformable convolutions -- we explore the use of these architectural designs as a defense in \secref{sec:defenses}.

\begin{figure}[ht]
\centering
\includegraphics[width=0.49\columnwidth]{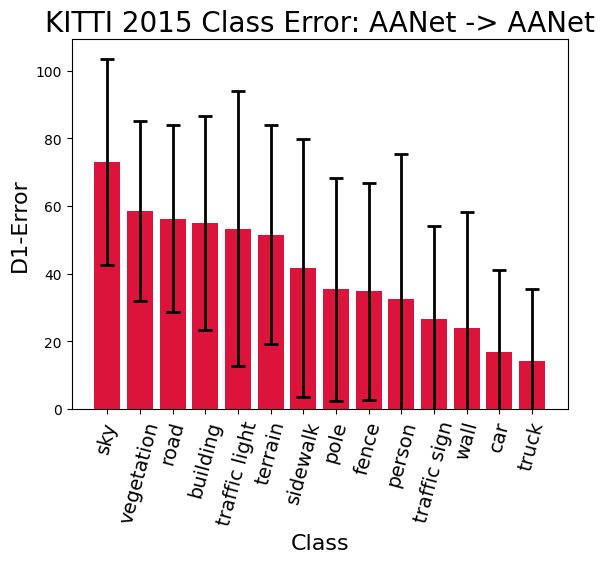}
\vspace{-1em}
\caption{\textit{D1-error for each semantic class for perturbed images}. Each class exhibit different levels of robustness. Homogeneous regions (sky, vegetation) are most vulnerable.}
\vspace{-0.5em}
\label{fig:aanet_class_error}
\end{figure}

\textbf{Effect on scene geometry.} To quantify how SUPs affect the estimated scene geometry, we compare the disparities estimated  for ``clean'' (no added perturbations, \figref{fig:aanet_disparity_distribution}-(a)) and perturbed (optimized on AANet, \figref{fig:aanet_disparity_distribution}-(b)) stereo pairs. \figref{fig:aanet_disparity_distribution} shows that the peak of the distribution shifts from $\approx$2 to $\approx$50px for all three networks. For PSMNet, we see an additional mode at $\approx$110px. Depth and disparity are inversely related, so the SUPs fool the network to predict the scene to be closer to the camera. We observe similar trends for DeepPruner and PSMNet (see Supp. Mat.).

\textbf{Robustness of semantic classes.} To analyze their effect on objects populating the scene, we use SDCNet \cite{zhu2019improving} to obtain segmentation maps for the KITTI 2015 validation set. We measure the per class error and found that different semantic classes exhibit different levels of robustness against adversaries. Specifically, \figref{fig:aanet_class_error} shows that \textit{sky} and \textit{vegetation} are the least robust with 72.96\% and 58.52\% D1-error, respectively; whereas, \textit{truck} (14.19\%) and \textit{car} (16.82\%) are the most robust. We observe that the least robust classes are largely homogeneous. We conjecture that these regions are most vulnerable because locally they give little to no information about scene structure, which leads to ambiguity when registering points between two images -- this is in contrast to sufficiently textured regions where unique correspondences are to be found. Thus, the network must rely on the regularizer (stored in the weights) learned from the training set to fill in the disparity for homogeneous regions.

\textbf{Effect on feature maps.} As DeepPruner and AANet use explicit matching to form their cost volume, there is a well-defined measure of data-fidelity to register the left and right images. So, to alter disparity, SUPs must corrupt the features used in the matching process. Hence, to quantify their effect, we measure the correlation between left and right feature maps before and after perturbing the images.

\begin{figure}[t]
\centering
\includegraphics[width=1.0\columnwidth]{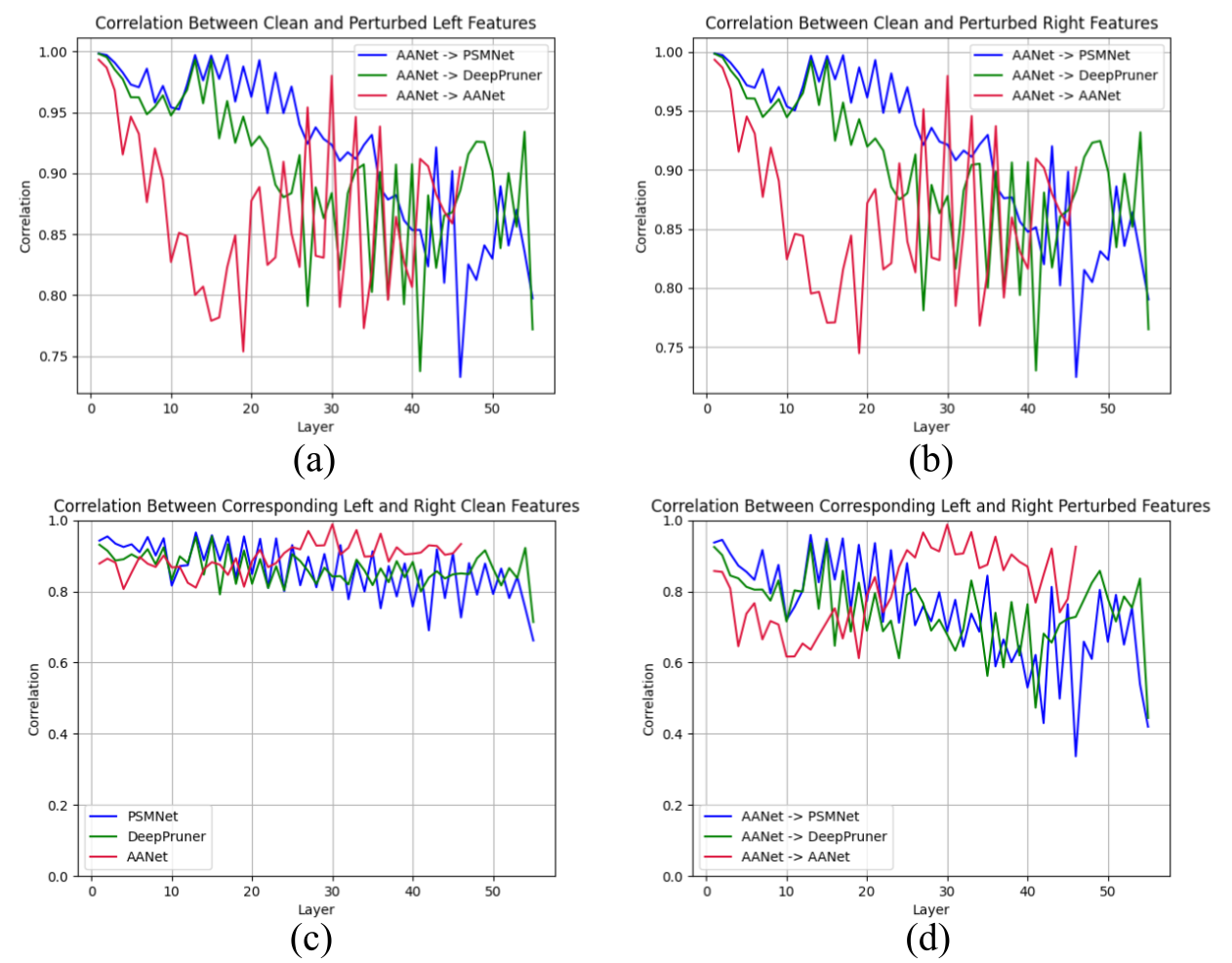}
\vspace{-2em}
\caption{\textit{Effect on features}. Clean and perturbed left (a) and~right (b) features grow uncorrelated. Features of clean stereo pairs are correlated (c), but after perturbation, become uncorrelated (d).}
\label{fig:aanet_feature_distance}
\vspace{-0.5em}
\end{figure}

\begin{figure*}[ht]
\centering
\includegraphics[width=0.99\textwidth]{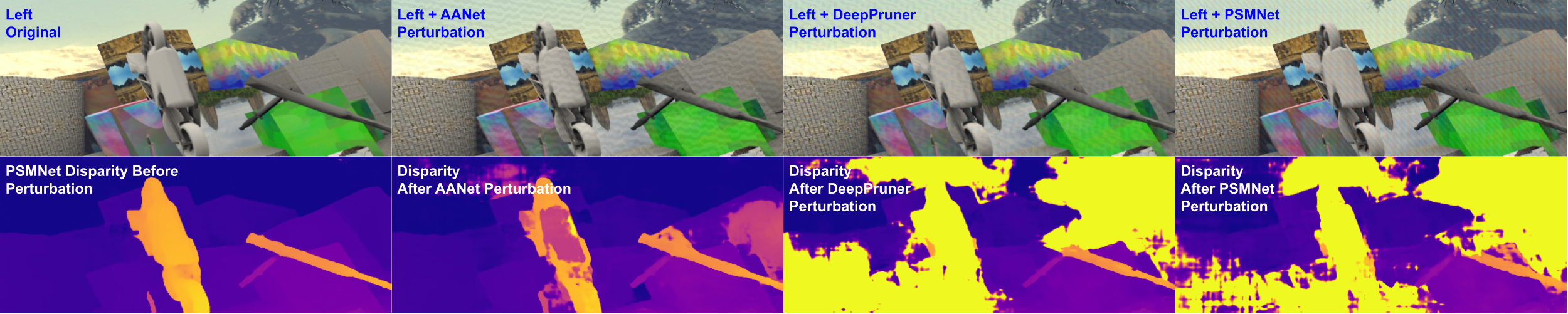}
\vspace{-0.8em}
\caption{\textit{Transferability to PSMNet}. Stereoscopic universal perturbations optimized on KITTI for AANet, DeepPruner, and PSMNet can generalize to stereo pairs in FlyingThings3D to corrupt the disparity estimation of PSMNet.}
\label{fig:transfer_to_psmnet}
\vspace{-1.2em}
\end{figure*}

\begin{figure*}[ht]
\centering
\includegraphics[width=0.8\textwidth]{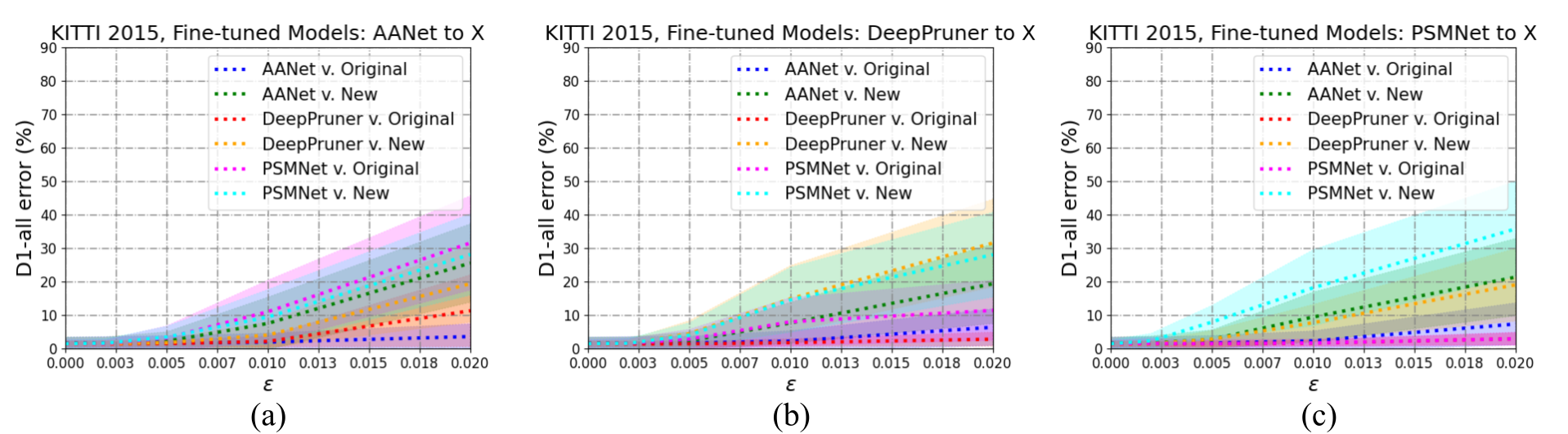}
\vspace{-0.8em}
\caption{\textit{Adversarial data augmentation}. AANet, DeepPruner, and PSMNet were fine-tuned with adversarial data augmentation. Each model was attacked with a perturbation trained for the original and fine-tuned AANet (a), DeepPruner (b), and PSMNet (c). Fine-tuning with adversarial data augmentation is an effective defense against SUPs trained for the original model, but does not fully mitigate a new adversary. Fine-tuning a model on SUPs optimized for it increases robustness against perturbations optimized for different architectures.}
\label{fig:finetuned_transfer}
\vspace{-1em}
\end{figure*}

Let $f_{\theta}^{(l)}$ be the $l$-th layer of the encoder shared between the stereo pair and $u \in \Omega$, the image domain. To quantify how SUPs corrupt the feature maps, we compute the correlation between $f_{\theta}^{(l)}(x_I(u))$ and $f_{\theta}^{(l)}(\hat{x}_I(u))$ for all $l$. \figref{fig:aanet_feature_distance}-(a, b) shows that when SUPs optimized for AANet are added to the input, the correlation between clean and perturbed left and right features grow uncorrelated from 1 to 0.76 during a forward pass i.e. the embedding function amplifies the effect of perturbation. We observe similar trends for DeepPruner and PSMNet (see Supp. Mat.).

While the observations in \figref{fig:aanet_feature_distance}-(a, b) may be sufficient to fool a classification network, i.e. the feature maps are transformed across a decision boundary, it is not sufficient for stereo matching. To fool a stereo network, SUPs must alter the correspondences between left and right image. In other words, for a pair of registered points $x_L(u)$ and $x_R(u-y_{gt}(u))$, where $y_{gt} \in \real^{H \times W}$ is the true disparity, the perturbations must cause the features of these similar points in the image to be dissimilar in embedding space. To quantify this, we first compute the correlation between the registered clean stereo pair $f_{\theta}^{(l)}(x_L(u))$ and $f_{\theta}^{(l)}(x_R(u-y_{gt}(u)))$ in \figref{fig:aanet_feature_distance}-(c). As expected the feature maps of the registered points are well correlated. In \figref{fig:aanet_feature_distance}-(d), we compute the correlation between the registered perturbed stereo pair $f_{\theta}^{(l)}(\hat{x}_L(u))$ and $f_{\theta}^{(l)}(\hat{x}_R(u-y_{gt}(u)))$. Indeed, the registered perturbed feature maps grow uncorrelated relative to the clean feature maps in the forward pass i.e. the perturbations cause similar regions in the RGB domain to be dissimilar in the embedding space, resulting in incorrect points being matched. Note that, like in \figref{fig:aanet_feature_distance}-(a) and \figref{fig:aanet_feature_distance}-(b), correlation between left and right AANet features increases from layer 20 to 30; this coincides with deformable convolutions. We conjecture that this may be related to AANet's relative robustness against adversaries.

\begin{figure*}[ht]
\centering
\includegraphics[width=1\textwidth]{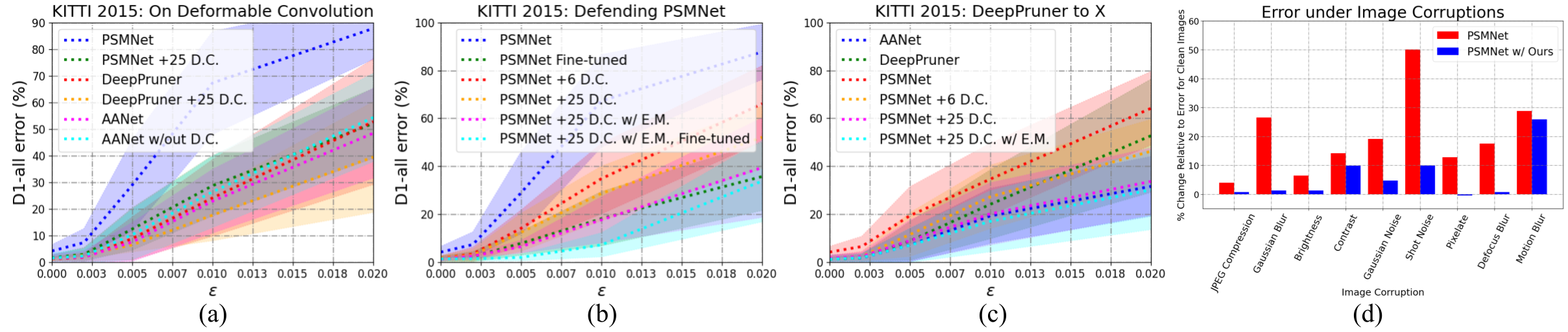}
\vspace{-2em}
\caption{\textit{Improving robustness against perturbations}. In (a, b), each variant was attacked by perturbations optimized specifically for them on image pairs. (a) Adding deformable convolution (DC) to PSMNet and DeepPruner improves their robustness, while removing it from AANet decreases its robustness. The least robust model, PSMNet, can achieve comparable performance to the most robust model, AANet, when using DC and explicit matching (EM). (b) Adding DC and EM to PSMNet achieves comparable results to adversarial training. (c) PSMNet with DC and EM is more robust than AANet under black-box attack, where the adversary was optimized for a different model (DeepPruner). (d) By applying our design principles to PSMNet, we improve its robustness to common image corruptions by $\approx$70\%.}
\label{fig:defenses}
\vspace{-1.2em}
\end{figure*}

\vspace{-0.2em}
\section{Towards Robust Deep Stereo Networks}
\vspace{-0.4em}
\label{sec:defenses}
\textbf{Adversarial data augmentation.} As shown in \cite{wong2020stereopagnosia}, fine-tuning with adversarial data augmentation is among the best performing defenses for stereo. 
Hence, we first fine-tuned each pretrained stereo model on KITTI 2015 with SUPs of $\epsilon \in \{0.002, 0.005, 0.01, 0.02\}$ trained for the model. The SUPs were randomly added the inputs with 50\% probability. In \figref{fig:finetuned_transfer}, we attack each fine-tuned model with a perturbation trained for the original and fine-tuned variant of each architecture.
\figref{fig:finetuned_transfer} shows that adversarial data augmentation improves the robustness of each model. When attacked by the SUPs it is fine-tuned on, AANet reduces in error from 48.43\% to 3.62\%, DeepPruner from 52.74\% to 2.83\%, and PSMNet from 87.72\% to 2.96\% for $\epsilon = 0.02$. 
New adversaries optimized for the fine-tuned networks are less effective, with AANet dropping to 25.54\% error, DeepPruner to 31.54\%, and PSMNet to 35.75\%.

\figref{fig:finetuned_transfer} also shows that fine-tuning a model on SUPs optimized for it increases robustness against SUPs optimized for different architectures.
For example, \figref{fig:finetuned_transfer}-(a) shows that fine-tuning reduces AANet in error from 48.43\% to 3.62\%, DeepPruner from 52.74\% to 11.32\%, and PSMNet from 87.72\% to 31.54\% when attacked by SUPs optimized for the original AANet with $\epsilon = 0.02$. Note that AANet has the lowest error against the original AANet adversary because it was fine-tuned on that perturbation, whereas DeepPruner and PSMNet are seeing it as a ``new'' adversary. Similar results are shown for SUPs optimized for DeepPruner (\figref{fig:finetuned_transfer}-(b)) and PSMNet (\figref{fig:finetuned_transfer}-(c)). We note that this process of optimizing SUPs and fine-tuning on them is time consuming, and the resulting networks are not fully robust.

\textbf{On explicit matching (EM) and deformable convolution (DC).} Instead, we propose to make simple modifications to the design of stereo networks. From our observations, EM increases robustness as PSMNet (no explicit matching) is more vulnerable than AANet and DeepPruner. \figref{fig:aanet_feature_distance} shows that the effect of SUPs is amplified by the embedding function, which ultimately fools the network; we conjecture that EM mitigates this by explicitly registering correspondences based on similarity rather than propagating the local signal. This intuition extends to DCs that learn convolutional offsets to regions locally similar to the element being convolved over and in effect ``avoids'' the adversarial signal -- \figref{fig:aanet_feature_distance} shows an increase in AANet's feature correlation that coincides with DCs. While the intent of DC is to minimize artifacts e.g. over-smoothing along occlusion boundaries by sampling features that are robust to local deformation and respect boundary conditions, we hypothesize that this filters out the perturbation signal that causes dissimilarities within a patch, and is the reason DC (and EM) allows AANet to be more robust. 

To assess DC as an inherent defense against adversarial perturbations, we trained (i) PSMNet and (ii) DeepPruner, each with 25 DCs, and (iii) AANet without DCs. We optimized six pairs of SUPs on KITTI for vanilla AANet, DeepPruner, PSMNet, and their variants. In \figref{fig:defenses}-(a), we show results on KITTI 2015, where each network was attacked by SUPs optimized specifically for them. We found that DCs do improve robustness as both DeepPruner and PSMNet produced lower D1-errors across all norms. For $\epsilon = 0.02$, DeepPruner drops in error from 52.74\% to 39.47\%, and PSMNet drops from 87.72\% to 52.10\%. Conversely, replacing DCs with regular convolutions can make a model more susceptible to adversaries -- AANet without DCs is less robust, as D1-error increase from 48.43\% to 54.32\%. In summary, simply designing a network with DC, the least robust model, PSMNet, can become comparable in performance to AANet, the most robust model.

Next, we assessed how EM and DC compare to adversarial training. We trained variants of PSMNet with (i) 6 DCs, (ii) 25 DCs, and (iii) 25 DCs and EM i.e. DeepPruner with 25 DCs. We performed adversarial fine-tuning on PSMNet and (iii). In \figref{fig:defenses}-(b), we observe for $\epsilon = 0.02$ that increasing the number of DCs and then adding explicit matching drops the D1-error of PSMNet from 87.72\% to 66.10\%, 52.10\%, and finally to 39.47\%. PSMNet fine-tuned on SUPs also performs well; however, with a D1-error of 35.75\%, it only marginally beats PSMNet with 25 DCs and explicit matching on $\epsilon = 0.02$. For all other norms, the two are comparable. The best performing variant is PSMNet with DCs and EM fine-tuned on SUPs, achieving D1-error of 33.85\%. So, simply using DCs and EM and following standard training protocols can yield more robust networks with no explicit intent for defense. Moreover, they can also be used in conjunction with existing defenses (i.e. adversarial fine-tuning) to yield even more robust networks.

In \figref{fig:defenses}-(c), we simulate the realistic black-box scenario where an attacker does not have access to a network (PSMNet or AANet) and crafts SUPs with an off-the-shelf model (DeepPruner). Replacing convolutions in PSMNet with DCs leads to immediate improvements in robustness with no loss in accuracy on clean images. With just 6 DCs, PSMNet becomes more robust than DeepPruner and with 25 it is on par with AANet. Incorporating PatchMatch into PSMNet (i.e. DeepPruner) with 25 DCs improves it to the most robust method. Note: fine-tuning on SUPs as data augmentation can further improve its robustness  (\figref{fig:defenses}-(b)).

Designing networks with DCs and inductive biases like EM not only improves robustness against SUPs, but also against common image corruptions i.e. lossy compression, blur and noise. \figref{fig:defenses}-(d) shows that PSMNet (red) is susceptible to blurring and shot noise where the latter can increase error by 50\%. Our design improves its robustness across all common corruption. Particularly, Gaussian and defocus blur, and pixelation have little effect -- we improve by as much as 80\% on shot noise and 70\% on average.

\vspace{-0.2em}
\section{Discussion}
\vspace{-0.4em}
\label{sec:discussion}
Stereoscopic universal perturbations (SUPs) exist and can generalize across architectures and datasets. SUPs can be partly mitigated by fine-tuning with adversarial data augmentation. However, doing so is costly in time and compute. Instead, we propose to address the robustness problem starting from the the design of deep networks. We have identified architectural elements, i.e. deformable convolutions and explicit matching, which can be easily incorporated into stereo networks with few lines of code and trained with standard protocol.  The resulting networks are comparable in robustness and performance to those without these elements, but fine-tuned on adversarial examples. Admittedly, SUPs do not exist in nature; nonetheless, our design is also applicable to common image corruptions. While the our scope is limited to stereo, many geometry problems i.e. optical flow share similar architectural designs. So we hope this work can contribute to robust systems in related fields.

\noindent\textbf{Acknowledgements.} We thank ARL W911NF-20-1-0158, ONR N00014-19-1-2229 and ARO W911NF-17-1-0304.

{\small
\bibliographystyle{cvpr}
\bibliography{cvpr}
}

\clearpage

\begin{center} 
    {\LARGE{\textbf{ \\ \vspace{1.0em}
    Supplementary Materials}}}
\end{center}

\vspace{1.0em}

\begin{appendices}

\begin{figure*}[ht]
\centering
\includegraphics[width=1\textwidth]{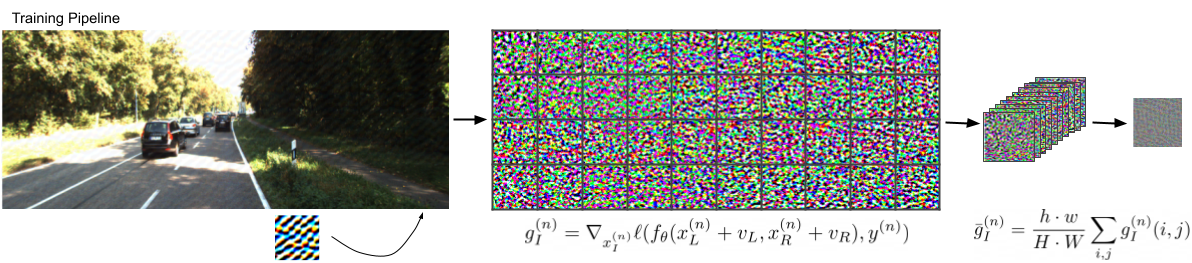}
\vspace{-0.0em}
\caption{\textit{Training pipeline}. Perturbations are tiled across their respective images. We take mean over the gradient with respect to each image for all tiles, which is used to update the each of the perturbations.}
\label{fig:training}
\vspace{-0.0em}
\end{figure*}

\section{Summary of Contents}
\label{sec:summary_of_contents}
We begin with a discussion on the existence of adversarial and universal perturbations for calibrated stereo in \secref{sec:existence_of_adversaries}.
In \secref{sec:training_pipline}, we illustrate our training pipeline in \figref{fig:training}.
In \secref{sec:implementation_details} we discuss implementation details, hyper-parameters, time and space requirements for training perturbations, fine-tuning with adversarial data augmentation, and retraining variants of AANet, DeepPruner, and PSMNet. We also describe the error metrics used throughout the paper. In \secref{sec:experiments_and_results}, we include results for KITTI 2012 (omitted in the main text) on the performance on each stereo network against our stereoscopic universal perturbations (SUPs). We also present additional analyses on the effect of SUPs on scene geometry, on robustness of semantic classes, and on correlation between clean and perturbed features extracted  for a given image and registered points between two images. In \secref{sec:deform_conv} we provide a discussion on the formulation of deformable convolutions and their use in stereo matching network. In \secref{sec:sample_efficiency}, we discuss the sensitivity of the proposed SUPs to number of training samples, and in \secref{sec:limitations}, the limitations of the SUPs as an attack and the proposed architectural designs to increase robustness of stereo networks against them. In \secref{sec:ethical_impact}, we discuss potential negative impact of our work and how we can mitigate them. Finally, in \secref{sec:qualitative_results}, we conclude with qualitative comparisons between perturbations crafted for the full image and the proposed $64 \times 64$ tiles. We also include additional qualitative results on the KITTI 2012, KITTI 2015 and FlyingThings3D datasets across all norms and transferability experiments.

\section{On the Existence of Adversaries for Stereo}
\label{sec:existence_of_adversaries}
Adversarial perturbations exist for a number of tasks from classification \cite{szegedy2013intriguing, goodfellow2014explaining}, object detection \cite{xie2017adversarial}, even single image depth prediction \cite{wong2020targeted}. While it may seem that they should exist for calibrated stereo, there is a qualitative difference between stereo and other single image based tasks where adversarial perturbations have been observed -- such are {\em purely inductive} tasks: Without training data and a strong inductive prior, a single image does not enable inference of depth or labels of objects. The likelihood is flat and adversarial perturbations have free reign to modify the outcome of inference, even to control the network to yield a desired 3D scene as outcome \cite{wong2020targeted}. Not so for stereo: binocular disparity is sufficient to infer depth wherever the image gradient is non-trivial, without any need for induction from a training set. One is not free to change the outcome of inference without observable changes in the likelihood. So, the fact that adversarial perturbations exist, i.e. Stereopagnosia \cite{wong2020stereopagnosia}, is indeed surprising for stereo, and that they would survive architectural changes even more so.

\section{Training Pipeline}
\label{sec:training_pipline}
In \figref{fig:training} we visualize the training pipeline for one iteration of our algorithm for crafting universal perturbations for stereo. Let $f_\theta(x_L, x_R) \in \real^{H \times W}$ be a pretrained stereo network that estimates the disparity between the left $x_L$ and right $x_R$ images of a stereo pair. Let $(v_L, v_R) \in \real^{h \times w}$ be a left and right perturbation subject to $h \mid H$ and $w \mid W$. First, we apply $(v_L, v_R)$ to $(x_L, x_R)$ over the entire image space, by evenly repeating the perturbation $v_I$ across $x_I$ with no overlap for $I \in \{L, R\}$. we then compute the gradient of the stereo network's loss $\ell(f_\theta(\cdot), y_{gt})$ with respect to each image $x_I$ in the stereo pair:
\begin{equation}
    g^{(n)}_I = \nabla_{x^{(n)}_{I}} \ell(f_\theta(\hat{x}^{(n)}_L, \hat{x}^{(n)}_R), y^{(n)}).
    \vspace{-0.2em}
\label{eqn:gradient_pipe}
\end{equation}
For a given stereo pair $(x_L, x_R)$, we take the mean over the gradient with respect to the image $g_I$ for all tiles:
\vspace{-0.2em}
\begin{equation}
    \bar{g}^{(n)}_I = \frac{h \cdot w}{H \cdot W}\sum_{i, j} g^{(n)}_I(i, j).
    \vspace{-0.2em}
\label{eqn:gradient_tile_pipe}
\end{equation}
This aggregated result can then be used to update the tile perturbation $v_I$ for $I \in \{L, R\}$.

\section{Implementation Details}
\label{sec:implementation_details}

\textbf{Datasets.} We optimize our perturbations on the KITTI dataset \cite{geiger2012we}, which contains $\approx$47K $376 \times 1240$ resolution stereo pairs of real-world outdoor driving scenarios. We evaluate them on the KITTI 2012, KITTI 2015 \cite{menze2015object}, and Scene Flow \cite{mayer2016large} stereo datasets using AANet \cite{xu2020aanet}, DeepPruner \cite{duggal2019deeppruner}, and PSMNet \cite{chang2018pyramid}. Due to computational limitations, we resize all images to $256 \times 640$ and adjusted disparity maps accordingly. Hence, the baseline error is slightly higher than those reported by each method.

KITTI 2012 contains 194 stereo pairs with sparse ground-truth disparities. KITTI 2015 contains 200 stereo pairs with high quality ground-truth disparity maps. Images in both datasets have dimension $376 \times 1240$. Following the KITTI validation protocol, KITTI 2012 is divided into 160 for training and 34 for validation and KITTI 2015 is divided into 160 for training and 40 for validation. We do not use any samples from KITTI 2012 or KITTI 2015 to optimize our perturbations, and only evaluate on the validation sets.

We demonstrate that our perturbations can generalize across datasets by testing them on Scene Flow \cite{mayer2016large} -- a synthetic dataset comprised of 35K training and 4370 testing $540 \times 960$ resolution images paired with ground-truth disparity maps. Like KITTI 2012 and 2015, we do not optimize our perturbations on Scene Flow and simply leverage their testing set, FlyingThings3D, in our evaluation.

We used PyTorch to implement our approach. We employed the publicly available code and pretrained models of AANet, DeepPruner, and PSMNet. We note that while AANet and PSMNet released separate pretrained models for KITTI 2012 and KITTI 2015, DeepPruner released one model trained on both datasets. All three stereo networks released pretrained models for Scene Flow. We note that the pretrained model on Scene Flow of PSMNet, provided by the authors, did not reproduce the results in their paper. We obtained D1-error $> 30\%$ and EPE $> 4$px when running on the Scene Flow test set. This is a known issue in their code repository. Hence, we fine-tuned the pretrained PSMNet model on Scene Flow, lowering the baseline D1-error to 5.57\% and the EPE to 1.27px, and used this model for our experiments. We note that DeepPruner provided two model variants, DeepPruner-Best and DeepPruner-Fast. Additionally, AANet provided AANet and AANet+. We used AANet and DeepPruner-Best for all of our experiments. To analyze the robustness of semantic classes, we used the implementation and pretrained model of SDCNet \cite{zhu2019improving}, a segmentation network for driving scenes.

\textbf{Hyper-parameters.} We considered the upper norms of $\epsilon \in \{0.002, 0.005, 0.01, 0.02\}$. We searched over learning rates of $\alpha \in \{0.00005, 0.0001, 0.0002, 0.0004, 0.0008\}$. We optimized SUPs on each network using square tiles of 16, 32, 64, and 128, and the full image size $256 \times 640$. As searching the full space of tile sizes would be intractable, we chose these as representative tiles at different scales. We omitted results for the $128 \times 128$ perturbation from the main paper due to space constraints, but describe them in \secref{sec:experiments_and_results}.

\textbf{Training perturbations.} We used an Nvidia GTX 1080Ti on a standard workstation for all of our experiments. To optimize SUPs, we iterated through the KITTI training set one time. Doing so took $\approx 5.5$hr to craft SUPs for AANet, $\approx 7.0$hr for DeepPruner, and $\approx 12.0$hr for PSMNet. Our procedure took $\approx 4.3$GB of GPU memory for AANet, $\approx 4.8$GB for DeepPruner, and $\approx 8.8$GB for PSMNet. As the SUPs are additive, they can be applied to an image in real time. We note that Alg. 1 uses zero initialization for the perturbations. When using the estimated disparity from the clean images as pseudo ground truth, this would yield no training signal. Thus, we added zero mean Gaussian noise to the pseudo ground truth.

\textbf{Fine-tuning stereo models with adversarial data augmentation.} To fine-tune the stereo models with adversarial data augmentation, we used 4 Nvidia GTX 1080Ti. The models were fine-tuned for 1000 epochs on KITTI 2015 with a batch size of 8. SUPs with upper norm $\epsilon \in \{0.002, 0.005, 0.01, 0.02\}$ were selected at random and added to the training images with 50\% probability. The learning rate was initially set to $1 \times 10^{-5}$, but switched to $5 \times 10^{-6}$ and $1 \times 10^{-6}$ after the 250th and 500th epoch, respectively. It took $\approx 9$hr, $\approx 11$hr, and $\approx 10$hr to fine-tune AANet, DeepPruner, and PSMNet, respectively. We note that on a standard workstation, this process can take up to a week to complete.

\begin{figure*}[ht]
\centering
\includegraphics[width=1.0\textwidth]{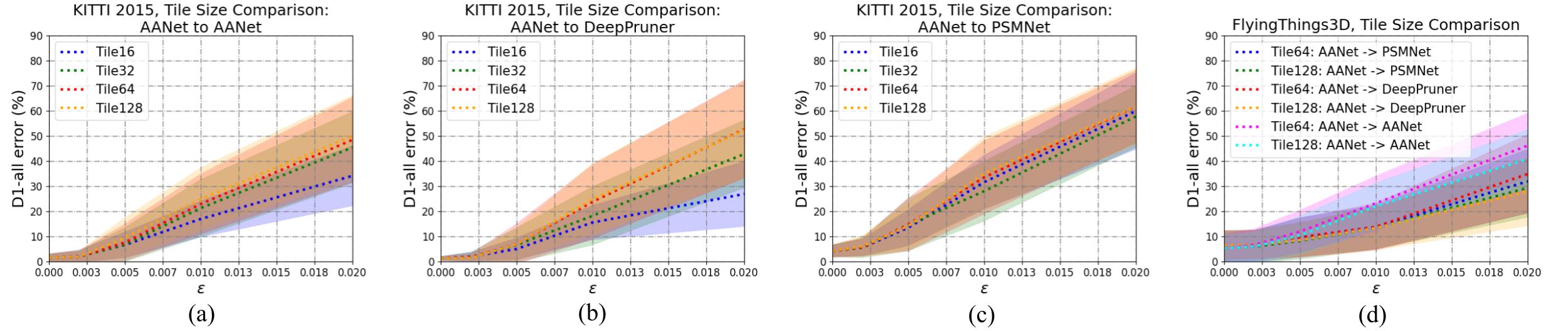}
\vspace{-1.5em}
\caption{\textit{Comparing tile sizes}. We optimized square tile perturbations with size 16, 32, 64, 128 on KITTI for AANet. (a, b, c) the smaller $16 \times 16$ and $32 \times 32$ tiles perform worst. (a) When applied to inputs from the dataset they are optimized on, the largest tile size $128 \times 128$ marginally outperforms $64 \times 64$. However, (d) shows that $64 \times 64$ consistently generalizes the best from KITTI to FlyingThings3D for all three architectures. There exists a trade-off between performance on the dataset on which a set of perturbations are optimized and their ability to transfer across models and datasets, where a smaller tile size e.g. $64 \times 64$ can generalize better, but a larger tile size e.g. $128 \times 128$ may deal more damage to the network and dataset for which the perturbations are optimized.}
\label{fig:tile_size_comparison}
\vspace{-0.5em}
\end{figure*}

\textbf{Training stereo models with deformable convolutions.} We trained a version of (i) PSMNet and (ii) DeepPruner from scratch, each with 25 deformable convolutions in the encoder. We also trained a version of (iii) PSMNet with 6 deformable convolutions, and a version of (iv) AANet without deformable convolutions. Four Nvidia GTX 1080Ti GPUs were used to train each variant, which took $\approx 4$ days per model. 

Both PSMNet models were trained with a batch size of 12, while DeepPruner was trained with a batch size of 16. PSMNet was trained on Scene Flow for 20 epochs, then fine-tuned on KITTI 2015 for 500 epochs. DeepPruner was first trained on Scene Flow for 64 epochs, then fine-tuned on a mixture of KITTI 2012 and KITTI 2015 for 1040 epochs. AANet was first trained on Scene Flow with a batch size of 22 for 64 epochs, fine-tuned on a mixture of KITTI 2012 and KITTI 2015 with a batch size of 12 for 1000 epochs, then fine-tuned on KITTI 2015 with a batch size of 8 for 1000 epochs. Note that DeepPruner is equivalent to PSMNet with an explicit matching module, so we sometimes refer to ``DeepPruner'' as ``PSMNet with explicit matching'' in our experiments.

\textbf{Evaluation metrics.} 
To evaluate the robustness of each stereo network, we use the official KITTI D1-error (the average number of erroneous pixels in terms of disparity) for KITTI 2012 and KITTI 2015 experiments:
\begin{align}
    \delta(i, j) &= |f_\theta(\cdot)(i, j) - y_{gt}(i, j)|, \\
    d(i, j) &= 
    \begin{cases} 
        1 & \mbox{if } \delta(i, j) > 3, \frac{\delta(i, j)}{y_{gt}(i, j)} > 5\%, \\
        0 & \mbox{otherwise} 
    \end{cases} \\
    \text{D1-error} &= \frac{1}{\|\Omega_{gt}\|} \sum_{i, j \in \Omega_{gt}} d(i, j),
    \label{eqn:d1-all-error}
\end{align}
and the official Scene Flow end-point-error (EPE) metrics on FlyingThings3D for generalization experiments:
\begin{align}
    \text{EPE} &= \frac{1}{\|\Omega_{gt}\|} \sum_{i, j \in \Omega_{gt}} \delta(i, j),
    \label{eqn:epe-error}
\end{align}
where $\Omega_{gt}$ is a subset of the image space $\Omega$ with valid ground-truth disparity annotations, $y_{gt} > 0$.

\section{Experiments and Results}
\label{sec:experiments_and_results}
In the main paper, we justified our use of a $64 \times 64$ sized perturbation for our experiments. Here we expand on our search of tile sizes. We omitted results on KITTI 2012 in the main paper due to space constraints, but present them in this section. Similarly, in the main paper, we only showed results on AANet for our experiments on the effect of SUPs on scene geometry, robustness of semantic classes, and effect on the feature extractor. Here, we also present results for SUPs trained for DeepPruner and PSMNet.

\textbf{Comparing tile sizes.} We optimized SUPs on KITTI for AANet at each tile size and attacked AANet (\figref{fig:tile_size_comparison}-(a)), DeepPruner (\figref{fig:tile_size_comparison}-(b)), and PSMNet (\figref{fig:tile_size_comparison}-(c)). The $16 \times 16$ and $32 \times 32$ tiles consistently performed worst, which justifies our choice not to explore smaller tiles. The $128 \times 128$ tile performed negligibly better than the $64 \times 64$ tile. However, \figref{fig:tile_size_comparison}-(d) shows that the $64 \times 64$ tile generalizes better than the $128 \times 128$ perturbation across all networks on FlyingThings3D. For SUPs with $\epsilon = 0.02$, $64 \times 64$ achieves 46.14\% error on AANet, 34.87\% on DeepPruner and 31.93\% on PSMNet, while $128 \times 128$ achieves 40.84\%, 27.89\%, and 29.33\% respectively. As demonstrated in the main paper, the full-size perturbations do worse than both, at 36.09\% error on AANet, 23.28\% on DeepPruner, and 25.35\% on PSMNet. We note that, in choosing the tile size, there is a clear trade-off between the performance on the model and dataset for which SUPs are optimized and the generalization to novel architectures and data distributions. For the purpose of universal perturbations that transfer across architectures and datasets, we choose the $64 \times 64$ tile size. However, given our results, we leave it up to the user to decide which tile size best suits their use case.

\begin{figure*}[ht]
\centering
\includegraphics[width=0.80\textwidth]{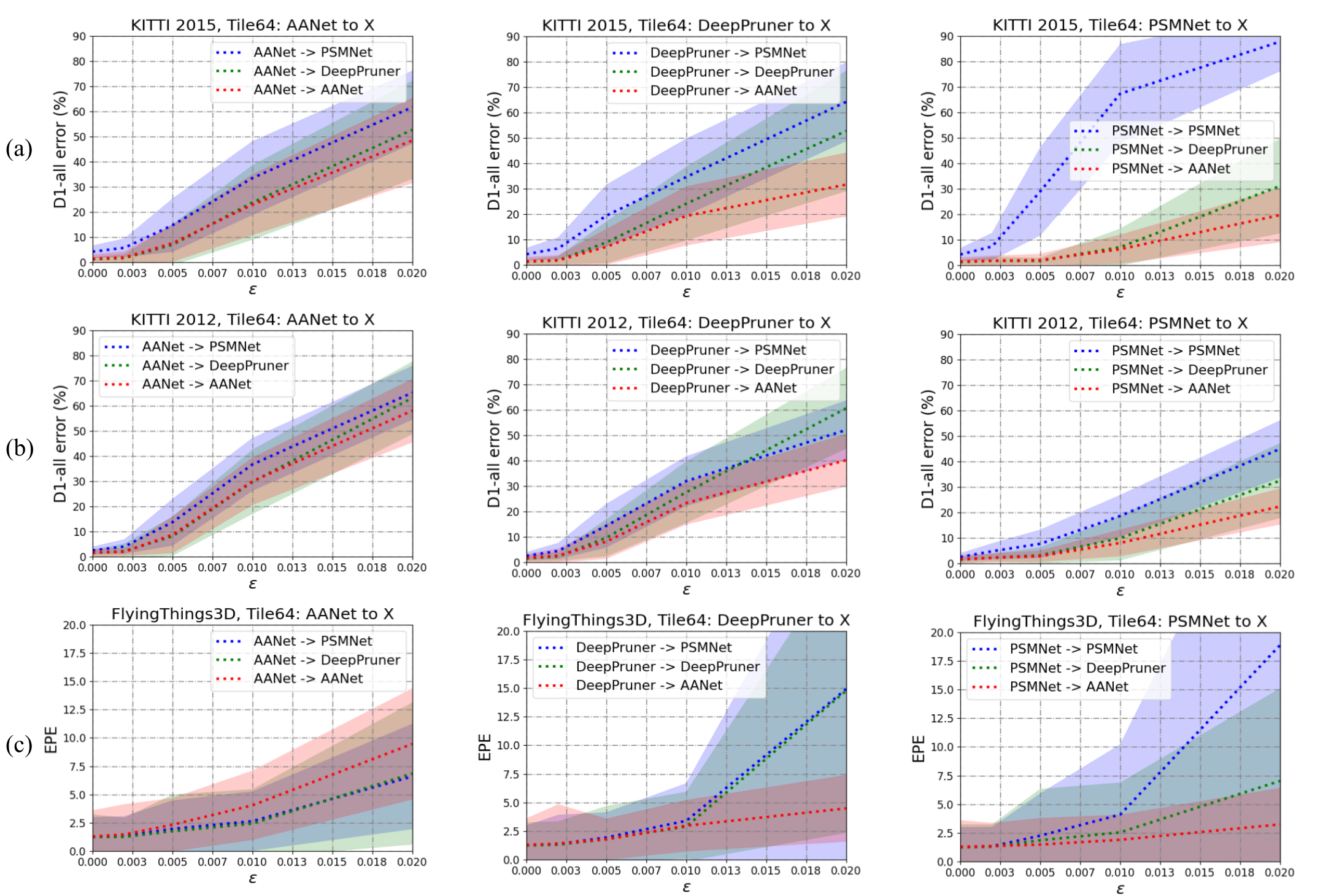}
\vspace{-0.5em}
\caption{\textit{Generalization across architectures and datasets}. We optimized stereoscopic universal perturbations on AANet, DeepPruner and PSMNet for KITTI (real dataset of outdoor driving scenario) and tested them on three datasets. We measure D1-error for KITTI 2012 and 2015, and EPE for FlyingThings3D. For all three networks, we add the perturbations optimized for each network to input stereo pairs from (a) KITTI 2015, (b) KITTI 2012, and (c) FlyingThings3D. The proposed universal perturbations optimized for a single network on KITTI can fool different network architectures across multiple datasets.}
\label{fig:attacking_quantitative_results_supp}
\vspace{-0.5em}
\end{figure*}

\begin{figure*}[ht]
\centering
\includegraphics[width=1.0\textwidth]{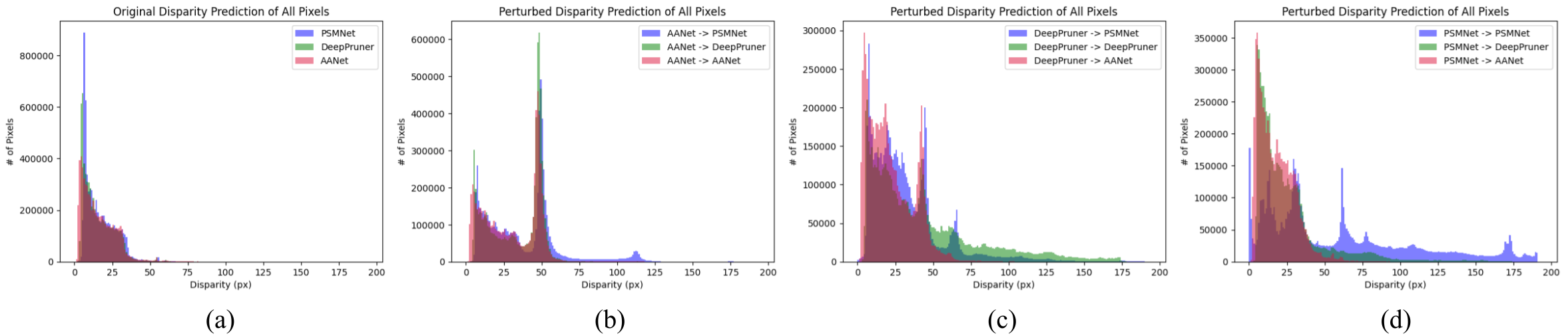}
\vspace{-1.5em}
\caption{\textit{Distribution of disparities before and after adding perturbations}. (a) Distribution of estimated disparities for AANet, DeepPruner and PSMNet on clean (no added perturbations) stereo pairs. Most disparities are concentrated at 2 pixels. Adding perturbations optimized for (b) AANet, (c) DeepPruner, and (d) PSMNet on KITTI to all three models. Disparities shift from 2 pixels to $\approx$50 pixels in (b), and to $\approx$40 and $\approx$60 pixels in (c) and (d). We note that in all cases, the disparities grow larger, meaning that the estimated depth grows smaller, so the perceived distances to objects populating the scene are closer than they should be.}
\label{fig:all_disparity_distribution}
\vspace{-0.5em}
\end{figure*}

\begin{figure*}[ht]
\centering
\includegraphics[width=0.70\textwidth]{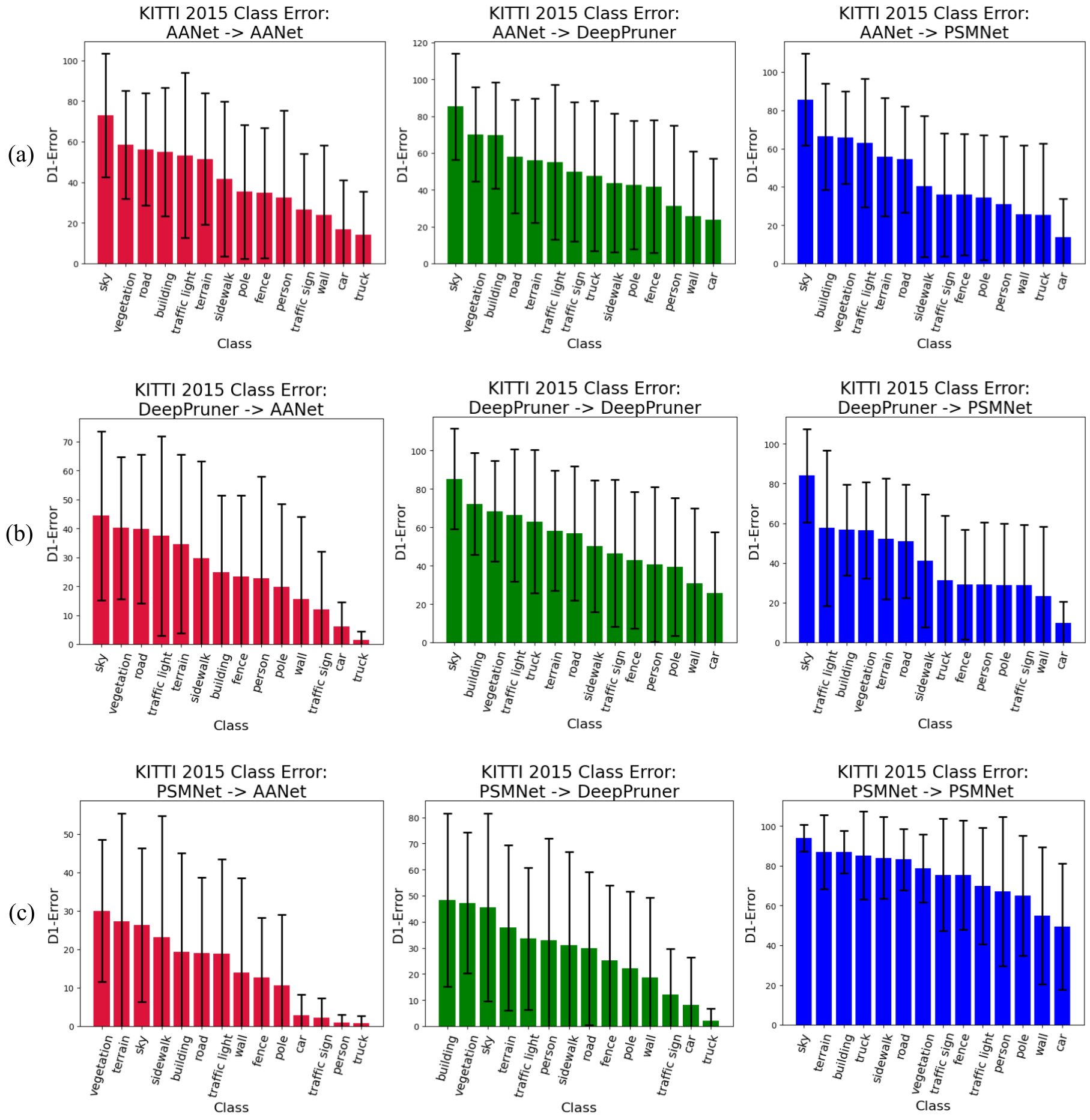}
\vspace{-0.0em}
\caption{\textit{Quantitative evaluation on KITTI 2015, D1-error for each semantic class}. We use SDCNet, an off-the-shelf semantic segmentation network to partition the image domain of stereo pairs from KITTI 2015 into semantic classes. We show the effect of perturbations optimized on KITTI on different classes for (a) AANet, (b) DeepPruner, and (c) PSMNet. Each semantic class exhibits a different level of robustness against adversaries. However, there are some common trends. For all perturbations and across all networks, the classes that are most susceptible are ``building'', ``vegetation'', ``sky'' and ``road''. The least susceptible are ``car'', ``person'', ``pole'' and ``traffic sign''.}
\label{fig:semantic_classes}
\vspace{-1em}
\end{figure*}

\textbf{Generalization across architectures and data.} We optimized three sets of $64 \times 64$ SUPs on KITTI for AANet, DeepPruner, and PSMNet, respectively. In \figref{fig:attacking_quantitative_results_supp}, we attack each network with each set of SUPs on three datasets: KITTI 2012 and 2015 (real datasets of outdoor driving scenarios), and Flyingthings3D (synthetic dataset of random ``flying'' objects). We report D1-error for KITTI 2012, and 2015, and EPE for FlyingThings3D. In the main paper, we omitted results for KITTI 2012 due to space constraints, but present them here.

In \figref{fig:attacking_quantitative_results_supp}-(a), we add the perturbations optimized for each network to the input stereo pairs from KITTI 2015 for all three networks. Here, KITTI 2015 is a held out data split from KITTI so the distribution of scenes follow that of the training set. When a set of perturbations are applied to the network for which they are optimized, as expected, they tend to be the most successful at corrupting the outputs of the network. We note that AANet (red lines) is consistently more robust against attacks and PSMNet (blue lines) is consistently more susceptible. 

In \figref{fig:attacking_quantitative_results_supp}-(b), we add the perturbations optimized for each network to input stereo pairs from KITTI 2012 for all three networks. Like KITTI 2015, KITTI 2012 is also a held out data split from KITTI, but contains different objects populating the scene. We observe similar trends as KITTI 2015, where AANet is still the most robust and PSMNet the least robust. We note that for all KITTI 2012 and 2015 experiments, all stereo models are trained on KITTI. 

In \figref{fig:attacking_quantitative_results_supp}-(c), we test the generalization of our perturbations across different data distributions. To this end, we add the perturbations optimized for each network to input stereo pairs from FlyingThings3D for all three networks. Here, FlyingThings3D is a synthetic dataset comprised of random ``flying'' objects where the scene distribution is differs greatly from that of KITTI. For this experiment, all of the stereo models are trained on Scene Flow datasets, which consists of Monkaa, Driving, and FlyingThings3D. Despite being optimized for a single network on KITTI, each pair of perturbations are able fool different network architectures trained on different datasets comprised of different scene distributions in a different domain. To the best of our knowledge, we are the first to demonstrate stereoscopic universal perturbations that generalize across architectures and data.

\begin{figure*}[ht]
\centering
\includegraphics[width=0.78\textwidth]{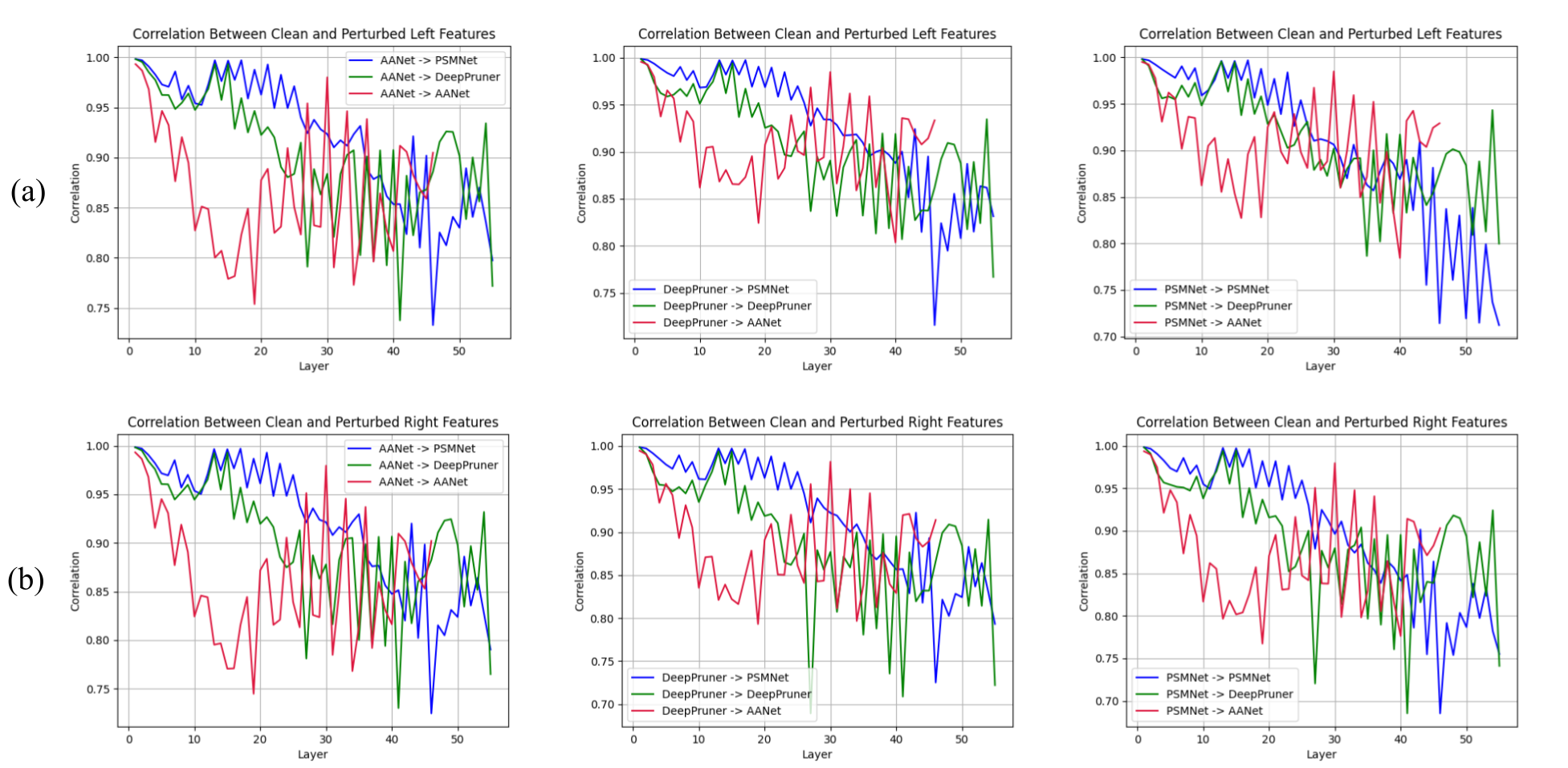}
\vspace{-0.5em}
\caption{\textit{Effect on clean and perturbed left and right feature maps}. Perturbations optimized for AANet, DeepPruner and PSMNet were added to KITTI 2015 stereo pairs. Correlation was computed between the (a) clean and perturbed left stereo images, and (b) clean and perturbed right stereo images. In both cases, the correlation decreased between the clean and perturbed feature maps. The perturbation signal is amplified by the encoding function as it is fed through the layers in a forward pass.}
\label{fig:feature_distance_clean_v_perturbed}
\vspace{-0.5em}
\end{figure*}

\begin{figure*}[ht]
\centering
\includegraphics[width=1.0\textwidth]{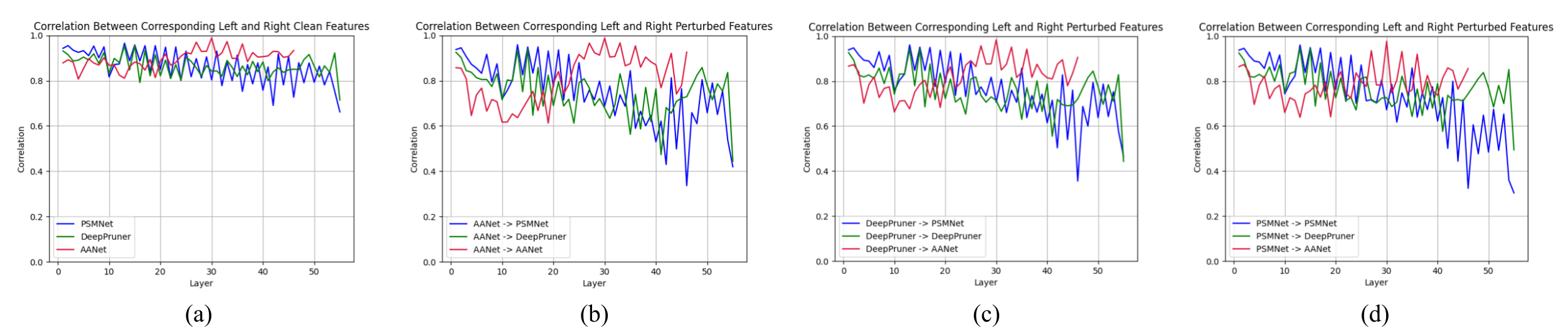}
\vspace{-1.5em}
\caption{\textit{Effect on corresponding left and right feature maps}. Perturbations optimized for AANet, DeepPruner and PSMNet were added to each stereo pair in KITTI 2015. (a) Correlation was computed between corresponding left and right clean features. Correlation was then computed between corresponding left and right perturbed features, using a perturbation optimized for AANet (b), DeepPruner (c), and PSMNet (d). The perturbed features become uncorrelated relative to the clean features. This suggests that universal perturbations cause similar regions in the RGB domain to be dissimilar in the embedding space, resulting in incorrect correspondences.}
\vspace{-0.5em}
\label{fig:feature_distance_warped}
\end{figure*}

\textbf{Effect on scene geometry.} In the main text, we showed that the mode of the distribution of disparities shifts from $\approx$2 pixels to $\approx$50 pixels when each network is attacked by SUPs optimized for AANet. To more effectively quantify how SUPs affect the predicted scene geometry, we plotted the distributions of disparity estimates for all networks tested, where clean (no added perturbation) baseline disparities are shown in \figref{fig:all_disparity_distribution}-(a) and disparities for perturbed stereo pairs are shown in \figref{fig:all_disparity_distribution}-(b,c,d). Specifically, we consider the change in disparity distribution for SUPs optimized and tested on the same network and SUPs optimized on other networks. In \figref{fig:all_disparity_distribution}-(b), we apply perturbations optimized for AANet, in \figref{fig:all_disparity_distribution}-(c) DeepPruner, and in \figref{fig:all_disparity_distribution}-(d) PSMNet. We note that all of these experiments were performed on the KITTI 2015 validation set.

As mentioned in the main text, there is a systemic increase in the estimated  disparities when perturbations are added to the input stereo pairs; in other words, a decrease in depth, where objects are perceived as closer than they really are. While the general trend is present for all networks, the effect is varied as SUPs optimized for AANet causes a sharp mode at $\approx$50 pixels whereas SUPs optimized for PSMNet and DeepPruner also create modes at $\approx$40 pixels and $\approx$60 pixels. While we do not conduct additional experiments to determine the reason for this bias, we hypothesize that it is induced by the dataset that the perturbations are trained on and will leave this analysis to future works. Interestingly, the same effect transfers to other datasets as well e.g. KITTI 2012 and FlyingThings3D, where visualizations of the output disparities are consistently closer than those estimated from clean images. Indeed, this shows that there are common vulnerabilities across images not only within a dataset, but across datasets and domains. We illustrate this phenomenon in Fig. 1, Fig. 3 and Fig. 8 in the main text, where larger disparities or smaller depths are depicted as brighter (yellow) regions in the colormap. We also observe a similar phenomenon in the additional qualitative results provided below (see \figref{fig:kitti2015_attack}, 
\figref{fig:kitti2012_attack}, 
\figref{fig:flyingthing3d_attack}, \figref{fig:kitti2015_to_psmnet}, \figref{fig:kitti2012_to_psmnet}, \figref{fig:flyingthing3d_to_psmnet}).

\textbf{Robustness of semantic classes.} In the main text, we observed that different semantic classes exhibit different levels of robustness against adversaries. We measured the per class error of the disparities estimated by AANet when attacked by a perturbation optimized for AANet. For completeness, we show the per class errors for each network when attacked by adversaries optimized for AANet (\figref{fig:semantic_classes}-(a)) DeepPruner (\figref{fig:semantic_classes}-(b)) and PSMNet (\figref{fig:semantic_classes}-(c)). To this end, we use SDCNet \cite{zhu2019improving}, an off-the-shelf semantic segmentation network to partition the image domain of stereo pairs from KITTI 2015 into semantic classes. We observe that each semantic class exhibits a different level of robustness against adversaries. However, there are some common trends across all the networks. \figref{fig:semantic_classes} shows that for all perturbations and across all networks, the classes that are most susceptible are ``building'', ``vegetation'', ``sky'' and ``road''. The least susceptible are ``car'', ``person'', ``pole'' and ``traffic sign''. We note that the most robust object classes tend to be those with rich textures. In contrast, the least robust classes are those that tend to be comprised of largely homogeneous regions e.g. ``sky'', ``road''. We hypothesize that this is due to the nature of the correspondence problem. Because stereo networks employ feature matching as a measure of data fidelity, regions with sufficiently exciting textures tend to have unique correspondences; whereas, there exists inherent ambiguity in locating correspondences in textureless, repeating patterns, and homogeneous regions. This leads to the network relying on the regularizer or the prior (learned from data) to fill in the gaps. The information of the regularizer is stored in the weights or parameters of a network via the training process. As the perturbation signal corrupts the activations of feature maps on which the weights operate on, it is thus corrupting the regularizer or prior, and so the prediction for textureless and homogeneous regions is worse.

\textbf{Effect on Feature Maps.}
\label{sec:feature_maps}
In the main text, we discussed the effect of stereoscopic universal perturbations (SUPs) on the encoder, or embedding function, for AANet. Here we show results for all three networks in \figref{fig:feature_distance_clean_v_perturbed}. We added the perturbations optimized for each network into stereo pairs from KITTI 2015 and forwarded them through each network. Similarly, we also fed clean stereo pairs without any added perturbations into each network. The per layer activations of the encoder, shared between the left and right images, were extracted for clean and perturbed stereo pairs. Correlation was then computed between the clean and perturbed left feature maps, and clean and perturbed right feature maps. \figref{fig:feature_distance_clean_v_perturbed}-(a) shows the correlation between clean and perturbed left image feature maps for each network, while \figref{fig:feature_distance_clean_v_perturbed}-(b) shows the same for the right image. In both cases, the correlation decreases between the clean and perturbed feature maps. This demonstrates that even though the input perturbation is constrained to be some $\epsilon$ small, the perturbation signal is amplified by the encoding function as it is fed through the layers in a forward pass. In effect, this can cause a network to output significantly different results.

However, stereo networks employ an explicit matching mechanism. Even though the perturbations may cause the clean and perturbed features to be different, as long as the true corresponding (registered) pixels between the left and right images are ``close'' in the embedding space, the correct disparity between the two images can be found. \figref{fig:feature_distance_warped} shows that indeed the perturbations not only causes clean and perturbed feature maps to be different, but also the corresponding registered left and right features to be different. Specifically, \figref{fig:feature_distance_warped}-(a) shows that registered clean features are well-correlated throughout all of the features maps in the encoder. This is expected. However, \figref{fig:feature_distance_warped}-(b, c, d) shows that left and right registered perturbed features become increasing uncorrelated relative to the clean features as they pass through the shared encoder. This suggests that our universal perturbations cause similar regions in the RGB domain to become dissimilar in the embedding space. As registered points in the images are no longer similar, this in turn fools the explicit matching modules to find incorrect correspondences, resulting in the wrong disparity estimates. We note that in both \figref{fig:feature_distance_clean_v_perturbed} and \figref{fig:feature_distance_warped}, the correlation of the features of AANet increase from layers 20 to 30. This phenomenon is present for all adversaries, and coincides with the use of deformable convolution in AANet. We conjecture that deformable convolutions may be related to the robustness of AANet. Hence, we use this observation to motivate the use of deformable convolutions (and explicit matching modules like correlation or PatchMatch) as part of our proposed architectural designs to increase robustness against SUPs (see Sec. 5 in the main paper).

\begin{figure}[t]
\centering
\includegraphics[width=1\columnwidth]{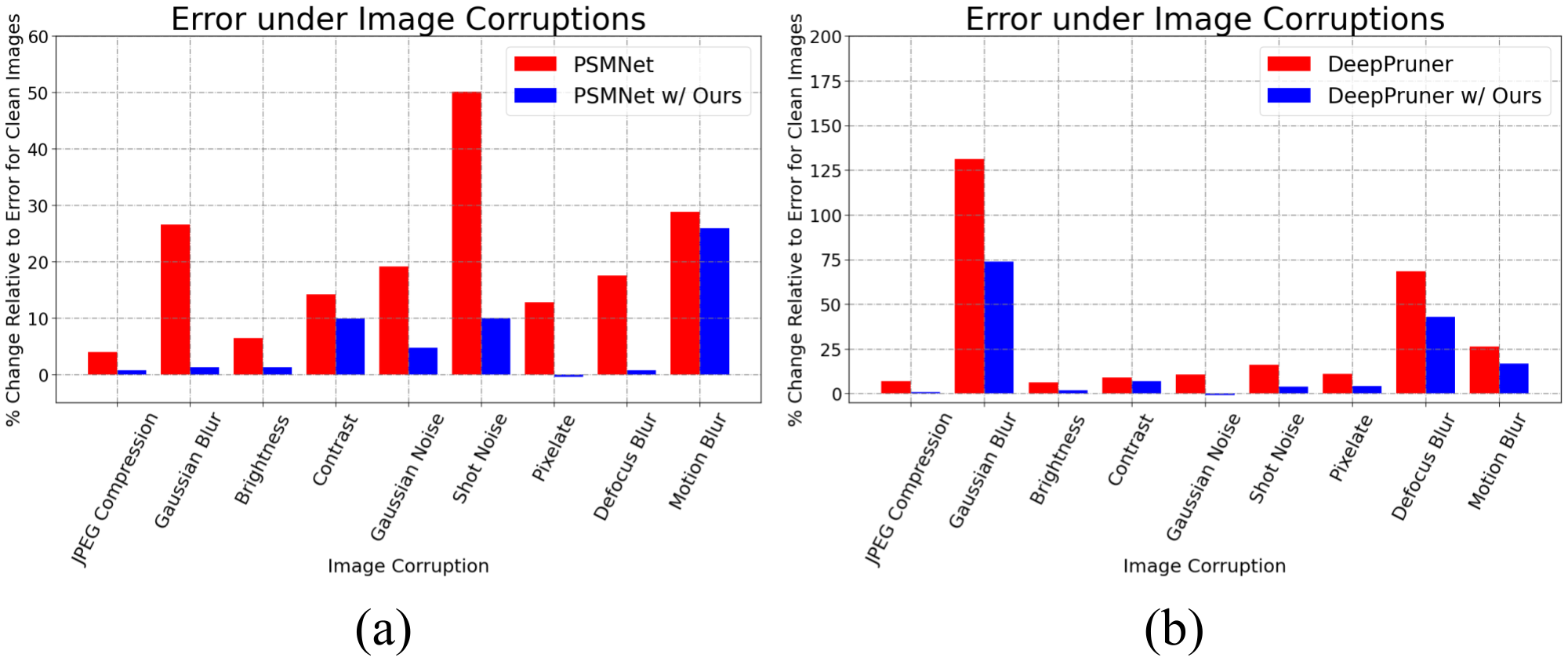}
\vspace{-1em}
\caption{\textit{Robustness to Image Perturbations}. (a) Robustness of PSMNet, PSMNet with 25 layers of deformable convolutions, (b) DeepPruner, and DeepPruner with 25 layers of deformable convolutions against different types of image corruptions.}
\label{fig:robustness_to_image_perturbations}
\vspace{-0.5em}
\end{figure}

\textbf{Robustness to Image Perturbations.} 
In the main text, we discussed the robustness of PSMNet and PSMNet with 25 DCs against common image corruptions i.e. lossy (JPEG) compression, Gaussian, defocus and motion blur, shot noise (Fig 10-(d)). As shown in \figref{fig:robustness_to_image_perturbations}-(a) and also in Fig 10-(d), our design improves over PSMNet by an average of 70\% to common image perturbations.   \figref{fig:robustness_to_image_perturbations}-(b) shows that DeepPruner (red), like PSMNet, is also susceptible to noise and blurring. In the case of Gaussian blur with a kernel size of 5 and a standard deviation of 1, the error increases by as much as 131\%. Our design of DeepPruner with 25 DCs (blue) is significantly more robust across all tested common image corruptions. There is an improvement of 43\% on gaussian blur and 37\% on defocus blur. Moreover, our design improves by 76\% on shot noise and 60\% on average.

\section{On Deformable Convolution}
\label{sec:deform_conv}
In the main paper, we found that replacing convolutional layers with deformable convolutional layers can increase the robustness of stereo networks. For the sake of completeness, in this section, we motivate the use of deformable convolutions in convolutional neural networks (CNNs) through their formulation. We then describe the use of deformable convolution in stereo matching networks.

\begin{figure}[t]
\centering
\includegraphics[width=1\columnwidth]{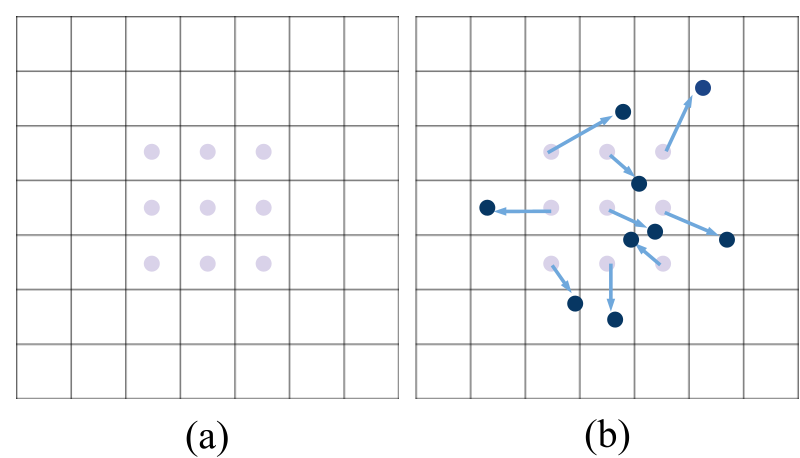}
\vspace{-1em}
\caption{\textit{Deformable convolution}. (a) The sampling grid of a standard $3 \times 3$ convolution. (b) The sampling locations of a deformable convolution (dark blue), adaptively offset from the regular sampling grid.}
\label{fig:deformable_convolution}
\vspace{-0.5em}
\end{figure}

\textbf{Motivation.} Due to rigid components, CNNs are limited at modeling geometric variations in object viewpoint, pose, scale, and part deformation. Conventional architectures are comprised of layers of fixed size convolutional filters over a regular grid with limited receptive field. While feature pyramids using spatial downsampling (e.g. max pooling and strided convolution) allow for modeling different scales of objects populating the scene, extensive geometric data augmentations are necessary to capture the variations listed above. Yet, it is not possible to fully capture all variations in the data. Additionally, the effective field of view is limited to a local neighborhood sampled regularly, which is ignorant of object boundaries. Hence, \cite{dai2017deformable, zhu2019deformable} proposed deformable convolutions to learn object deformations by directly conditioning on the data. Deformable convolution adds 2D offsets, predicted based on the input feature map, to the sampling grid of standard convolution. It allows the sampling grid to deform freely and hence capture visibility phenomena such as occlusions, as visualized in \figref{fig:deformable_convolution}.

\textbf{Formulation.} Let $\mathbf{x}$ be an input feature map and $\mathbf{y}$ be an output feature map. The standard 2D convolution is a two-step operation: 1) sample $\mathbf{x}$ over a regular grid $\mathcal{R}$; 2) compute a weighted sum of the sampled values to generate $\mathbf{y}$. Formally, for each location $\mathbf{p}_0$ in $\mathbf{y}$,
\begin{equation}
    \mathbf{y}(\mathbf{p}_0) = 
    \sum_{\mathbf{p}_n \in \mathcal{R}} \mathbf{w}(\mathbf{p}_n) \cdot \mathbf{x}(\mathbf{p}_0 + 
    \mathbf{p}_n)
\label{eqn:2d_convolution}
\end{equation}
where $\mathbf{p}_n$ are the locations in $\mathcal{R}$.

In deformable convolution, the sampling grid $\mathcal{R}$ are augmented with offsets 
$\{\Delta \mathbf{p}_n | n = 1, ..., |\mathcal{R}|\}$. \eqnref{eqn:2d_convolution} extends to
\begin{equation}
    \mathbf{y}(\mathbf{p}_0) = 
    \sum_{\mathbf{p}_n \in \mathcal{R}} \mathbf{w}(\mathbf{p}_n) \cdot \mathbf{x}(\mathbf{p}_0 + 
    \mathbf{p}_n + 
    \Delta \mathbf{p}_n).
\label{eqn:2d_deformable_convolution}
\end{equation}
Here, we sample at offset locations $\mathbf{p}_n + \Delta \mathbf{p}_n$. Because the offset $\mathbf{p}_n$ can be fractional,  $\mathbf{x}(\mathbf{p}_0 + \mathbf{p}_n + \Delta \mathbf{p}_n)$ may not correspond to an actual element and is subject to quantization effects. Hence, $\mathbf{x}(\mathbf{p}_0 + \mathbf{p}_n + \Delta \mathbf{p}_n)$ is computed via bilinear interpolation.

\textbf{Use in stereo matching.} AANet \cite{xu2020aanet} introduce an intra-scale aggregation (ISA) module for cost aggregation, designed to adaptively sample points from regions of similar disparity. The authors intuit that adaptive sampling prevents the edge-fattening issue at disparity discontinuities \cite{scharstein2002taxonomy}. Formally, let $\mathbf{C} \in \mathbb{R}^{D \times H \times W}$ be a cost volume with maximum disparity D, height H, and width W. For $K^2$ sampling points, the ISA computation is
\begin{equation}
    \widetilde{\mathbf{C}}(d, \mathbf{p}) =
    \sum^{K^2}_{k = 1} w_k \cdot \mathbf{C}(d, \mathbf{p} + \mathbf{p}_k + \Delta \mathbf{p}_k)
\label{eqn:isa_module}
\end{equation}
where $\widetilde{\mathbf{C}}(d, \mathbf{p})$ is the aggregated cost at pixel $\mathbf{p}$ for disparity $d$, $w_k$ is the weight for point $k$, $\mathbf{p}_k$ is a fixed offset from $\mathbf{p}$, and $\Delta \mathbf{p}_k$ is a learned offset. Since the formulations are similar, ISA is implemented with deformable convolution. We note that in addition to the ISA module, AANet also employs deformable convolutions in its encoder to reduce sampling across object boundaries, which minimizes the bleeding effect often seen in backprojected point clouds. While the proposed use case is mainly to handle object deformation and variations in view point, we found that deformable convolutions are amicable towards defending against adversaries as their formulation naturally allows for ``avoiding'' certain signals e.g. occlusion boundaries, part deformation, and perhaps adversarial perturbations present in the input.

\begin{figure}[t]
\centering
\includegraphics[width=0.70\columnwidth]{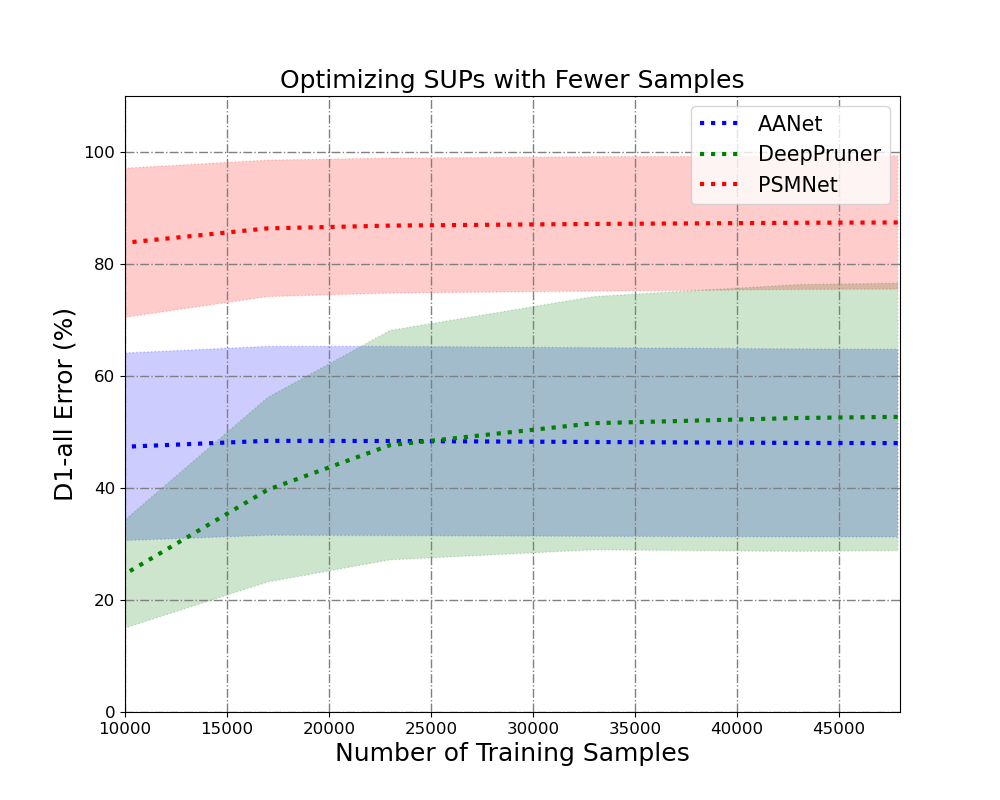}
\caption{\textit{Sample efficiency}. We optimize SUPs for AANet, DeepPruner and PSMNet with fewer samples and test their effectiveness on KITTI 2015. SUPs for AANet and PSMNet be still be effective even when optimized with 79\% fewer samples; the effectiveness of SUPs optimized for DeepPruner start to decrease when optimized with less than 60\% of the dataset.}
\label{fig:sample_efficiency}
\end{figure}

\section{On Sample Efficiency}
\label{sec:sample_efficiency}
In Fig 5, 8 of main text, SUPs were optimized using KITTI (outdoor driving, real domain) and are added to images from the FlyingThings3D test set (randomly generated scenes, synthetic), part of the Scene Flow \cite{menze2015object} datasets. These perturbed images are fed to AANet \cite{xu2020aanet}, DeepPruner \cite{duggal2019deeppruner} and PSMNet \cite{chang2018pyramid} that were trained on Scene Flow. For this particular set of experiments, our SUPs were trained with substituted data from KITTI and zero-shot transferred to FlyingThings3D.

In this section, we study the sample efficiency of SUPs. We optimized SUPS for AANet, DeepPruner and PSMNet using the KITTI raw dataset for varying training sample sizes. \figref{fig:sample_efficiency} shows that SUPs for AANet and PSMNet are still effective when optimized with just 21\% of KITTI; whereas SUPs for DeepPruner start to decrease in effectiveness when we remove 60\% of the samples.

\section{Limitations}
\label{sec:limitations}
While we have demonstrated stereoscopic universal perturbations (SUPs) that generalize across network architectures, datasets and domains and showed that deformable convolutions and explicit matching can help mitigate them, there exist limitations on both fronts. Despite reaching as high as 87\% in D1-error error on PSMNet, SUPs has a limited effectiveness on AANet and reaches 50\% D1-error -- this is equivalent to half of the perceived scene being corrupted. In none of our experiments do the error reach 100\%, meaning that there are still portions of the scene that are reasonably correct. 

We note that such perturbations are not undefendable; we showed that fine-tuning with adversarial data augmentation does mitigate them to some extent, but not fully. For example D1-error decreased from 87.72\% to 2.96\% for PSMNet when attacked by the adversary it was trained on, but when attacked by new adversaries, it still has a 35.75\% error. While our proposed architecture designs involving deformable convolutions and explicit matching does help, they also do not fully mitigate the attack. 

Our scope is limited to stereo \cite{chang2018pyramid,duggal2019deeppruner, poggi2020self,xu2020aanet}; however, the component proposed in our architectural design. i.e. deformable convolutions and inductive biases like explicit matching, are used in many geometric problems such as optical flow \cite{aleotti2020learning,lao2017minimum,lao2018extending,lao2019minimum, sun2018pwc,teed2020raft,yang2018conditional}, multi-view stereo \cite{chen2019point,gu2020cascade,yao2018mvsnet,yao2019recurrent,wang2021patchmatchnet}, monocular depth prediction \cite{fei2019geo,godard2019digging,poggi2020uncertainty,poggi2022real,ranftl2021vision,watson2019self,wong2019bilateral}, and depth completion \cite{lin2022dynamic,hu2021penet,park2020non,yang2019dense,wong2020unsupervised,wong2021adaptive,wong2021learning,wong2021unsupervised,zhu2021robust}. These are problems where adversarial and universal perturbations are less studied. So there is a long road ahead, but we hope that our findings will be useful towards realizing robustness deep neural networks. 

\section{Discussion of Potential Negative Impact}
\label{sec:ethical_impact}
Deep learning models have been extensively deployed for many applications. Hence, adversarial perturbations have been treated as a security concern. These concerns were initially far fetched, as adversarial perturbations were optimized per-image instance; it would be computationally infeasible to corrupt a model in real-time. Yet, the discovery of universal adversarial perturbations made the security threat more realistic.

We showed that stereoscopic universal perturbations (SUPs) that generalize across architecture and data exist. These can be applied effectively to attack models in the black-box setting, and so appear to present a more immediate security threat. Yet, we believe these perturbations will not cause damage outside of the academic setting. While our perturbations can be realized as a filter to be placed on top of a camera lens, autonomous agents typically have a myriad other sensors. An autonomous system should not fail due to a corrupted disparity map as long as it can rely on sensor measurements from other sources.

Rather, we have used stereoscopic universal perturbations to better understand the robustness of stereo networks. By identifying how SUPs can corrupt stereo networks, we were able to motivate several architectural designs (see Sec. 5, main text) that ultimately improve the robustness of stereo models. Adversarial perturbations expose inherent problems with our deep networks -- yet, we view them as an opportunity to unravel our black-box models and develop more robust representations.

Nonetheless, to mitigate them, one can redesign models with the proposed architectural changes, leverage adversarial data augmentation to fine-tune existing models, or utilize techniques to denoise, purify, or rectify the input as discussed in our Related Works section (Sec. 2, main text). Additionally, we will also restrict code and perturbations usage via our release license. 

\section{Additional Qualitative Results}
\label{sec:qualitative_results}
\begin{figure*}[ht]
\centering
\includegraphics[width=1.0\textwidth]{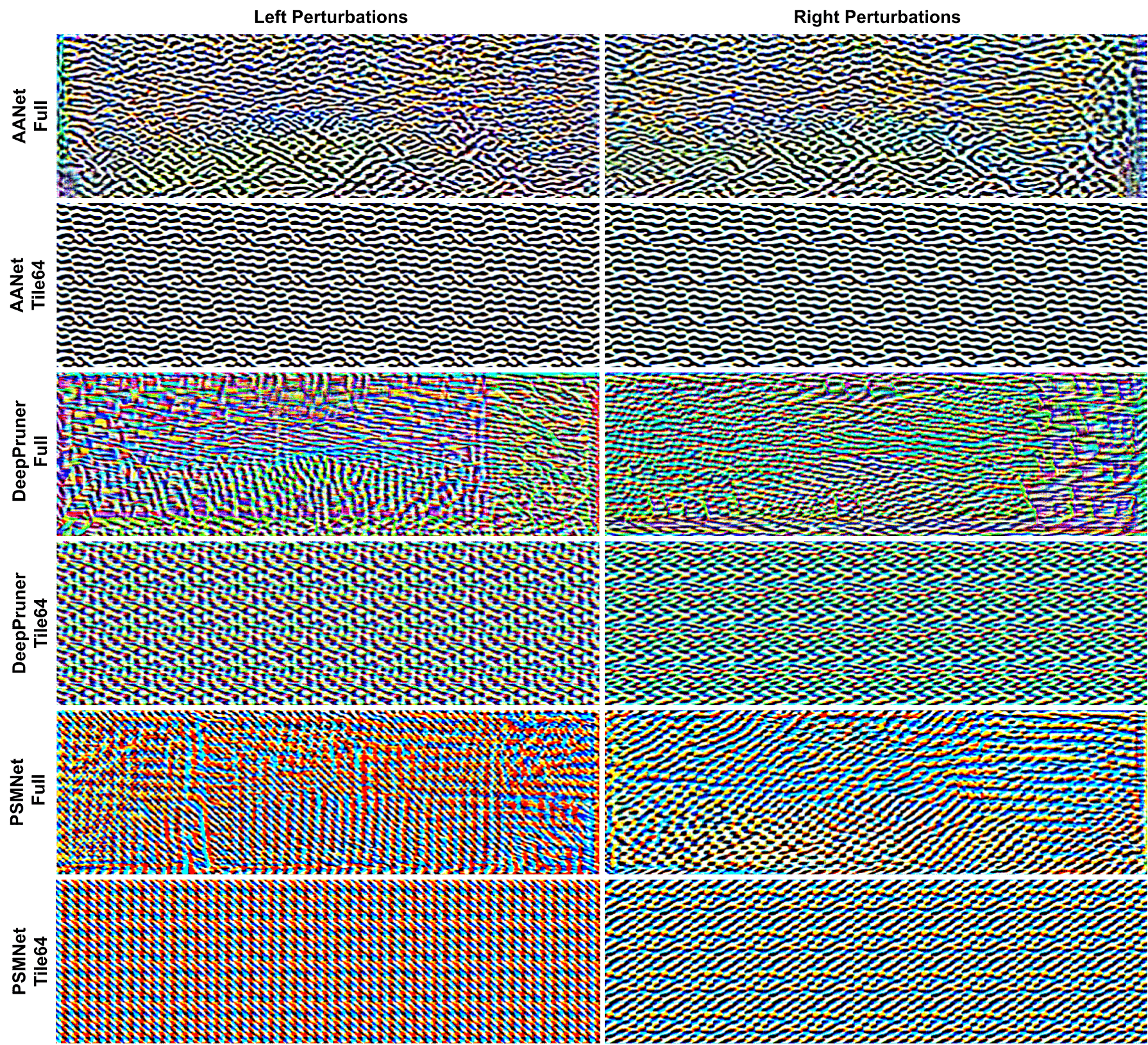}
\caption{\textit{Examples of Stereoscopic universal perturbations} (SUPs) optimized for AANet, DeepPruner, and PSMNet on the KITTI dataset. Each panel of two rows shows SUPs optimized over the full $256 \times 640$ image and the $64 \times 64$ sized perturbations, tiled across the image domain, for each method. We note that for all full image size SUPs, there are structural artifacts biased by the scenes in the dataset, which limits their generalization capabilities.}
\label{fig:all_perturbations}
\end{figure*}

In \figref{fig:all_perturbations} we show SUPs optimized for AANet, DeepPruner, and PSMNet on the KITTI dataset. Each panel of two rows shows SUPs optimized over the full $256 \times 640$ image and the $64 \times 64$ sized perturbations, tiled across the image domain, for each method. The tiled perturbations are then used for each subsequent visualization. We note that for all full image size SUPs, there are structural artifacts biased by the scenes in the dataset. For instance, full image size SUPs for PSMNet shows a road pattern in the both perturbation images. As a result, these dataset specific structures limit the generalization of SUPs that were optimized over the full image. Unlike them, $64 \times 64$ tiles do not contain any of these structures as they are spatially invariant, enabling them to transfer across datasets and domains.

\begin{figure*}[ht]
\centering
\includegraphics[width=1.0\textwidth]{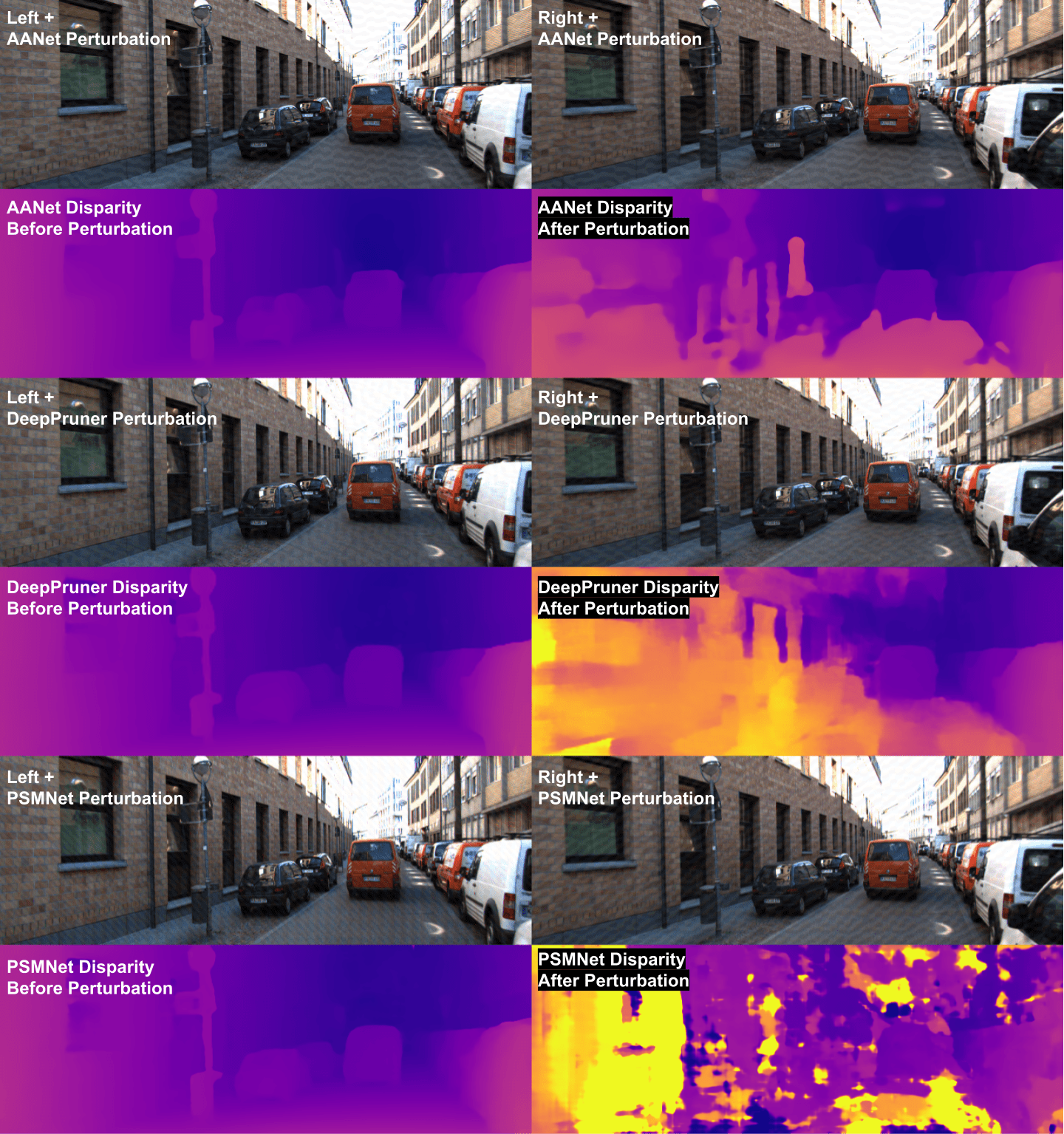}
\vspace{-1em}
\caption{\textit{Attacking AANet, DeepPruner, and PSMNet on a scene from KITTI 2015 using the SUP trained for each model}.}
\vspace{-1em}
\label{fig:kitti2015_attack}
\end{figure*}

\begin{figure*}[ht]
\centering
\includegraphics[width=1.0\textwidth]{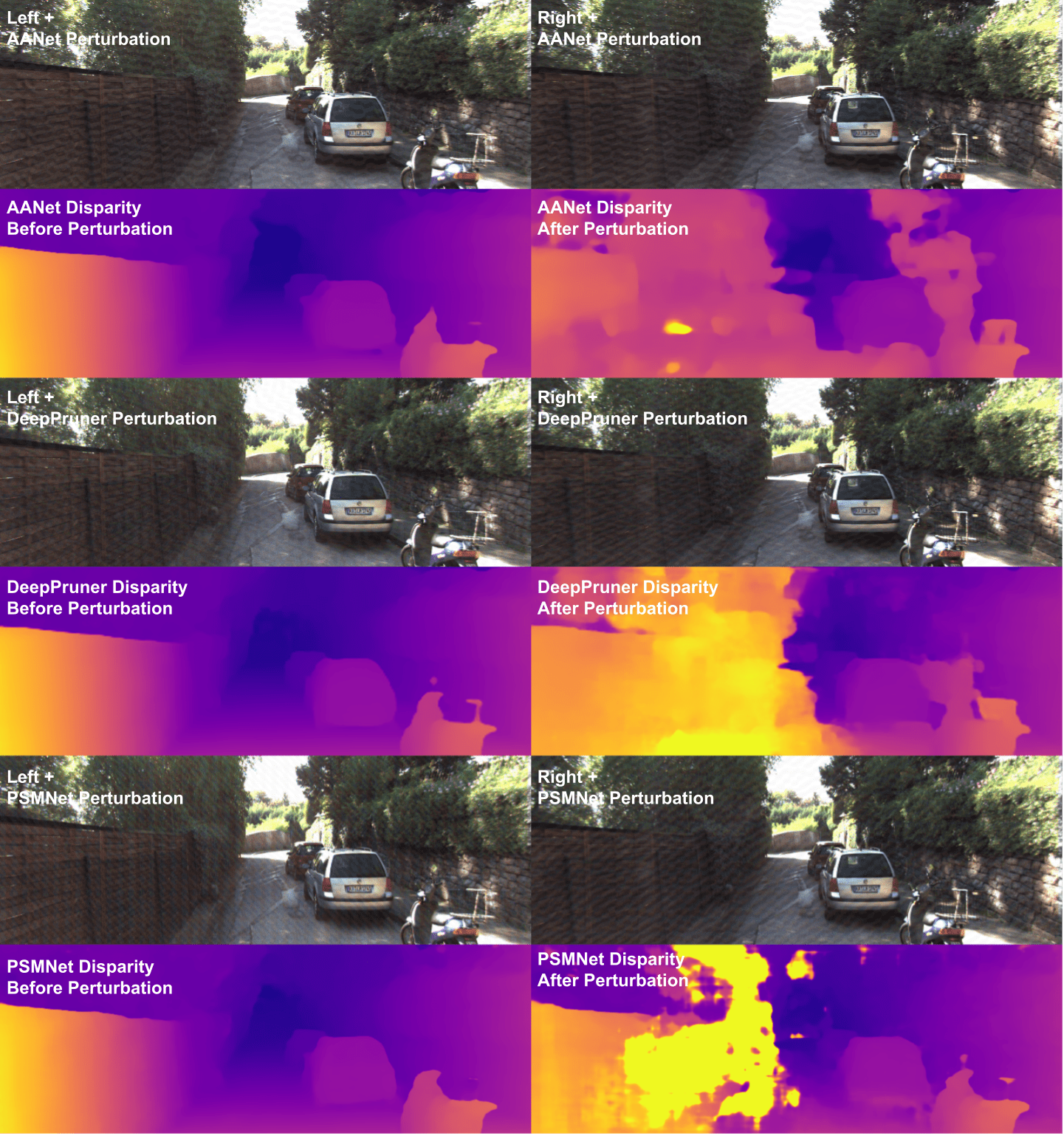}
\vspace{-1em}
\caption{\textit{Attacking AANet, DeepPruner, and PSMNet on a scene from KITTI 2012 using the SUP trained for each model}.}
\vspace{-1em}
\label{fig:kitti2012_attack}
\end{figure*}

\begin{figure*}[ht]
\centering
\includegraphics[width=1.0\textwidth]{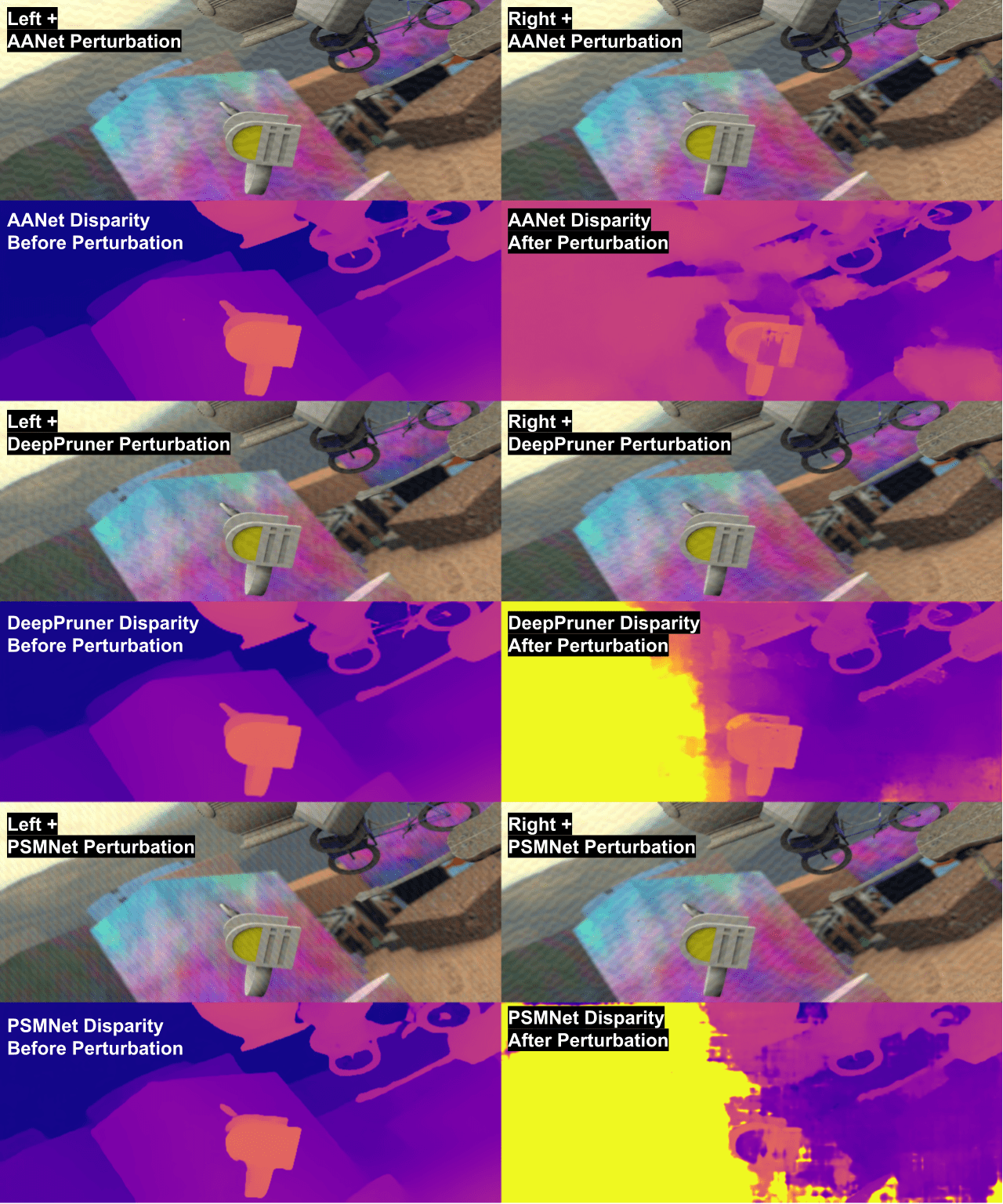}
\caption{\textit{Attacking AANet, DeepPruner, and PSMNet on a scene from FlyingThings3D using the SUP trained for each model}.}
\label{fig:flyingthing3d_attack}
\end{figure*}

\begin{figure*}[ht]
\centering
\includegraphics[width=1.0\textwidth]{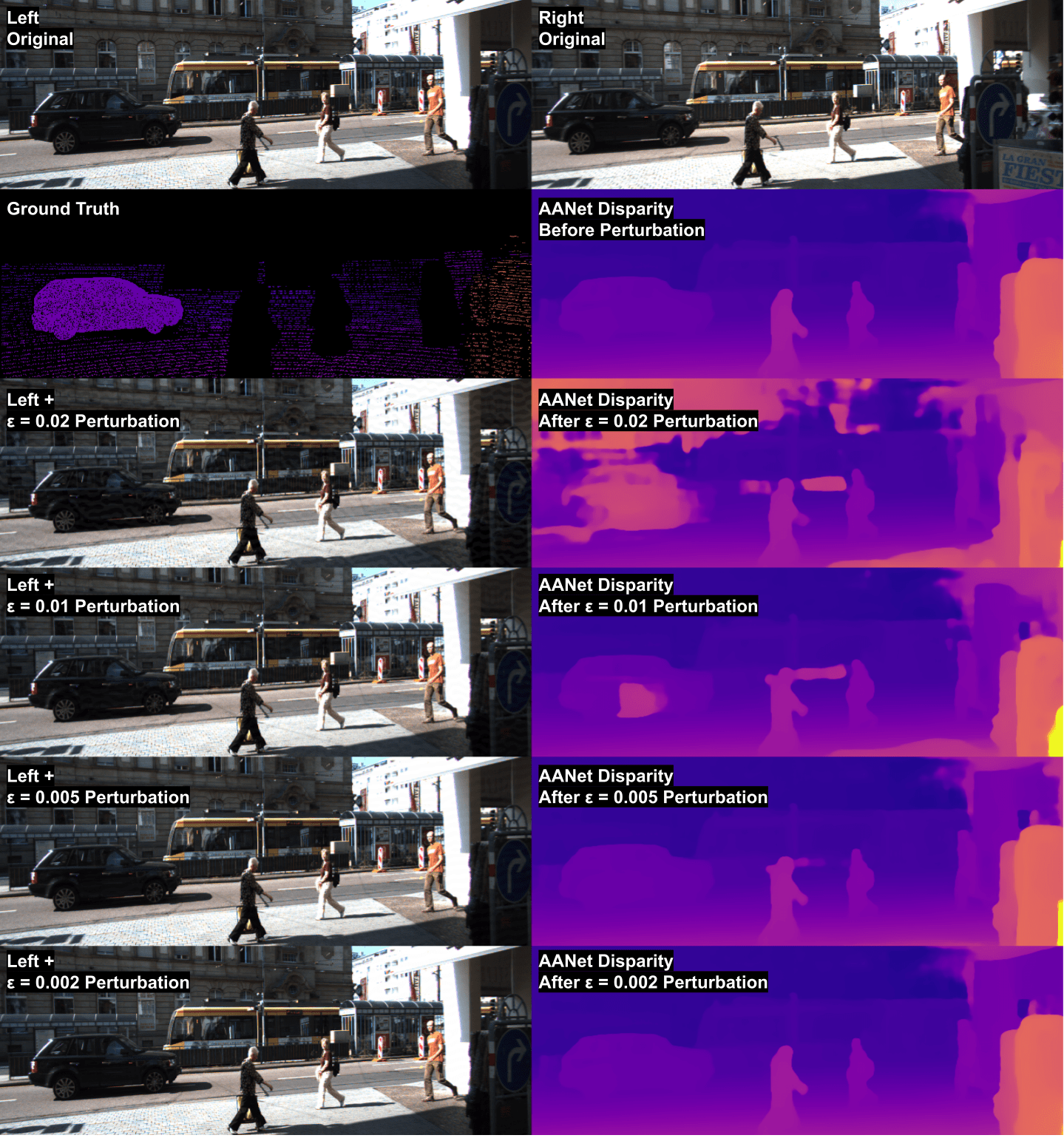}
\caption{\textit{Attacking AANet at different upper norms $\epsilon \in \{0.002, 0.005, 0.01, 0.02\}$ for a scene from KITTI 2015}.}
\label{fig:aanet_kitti2015_different_norms}
\end{figure*}

\begin{figure*}[ht]
\centering
\includegraphics[width=1.0\textwidth]{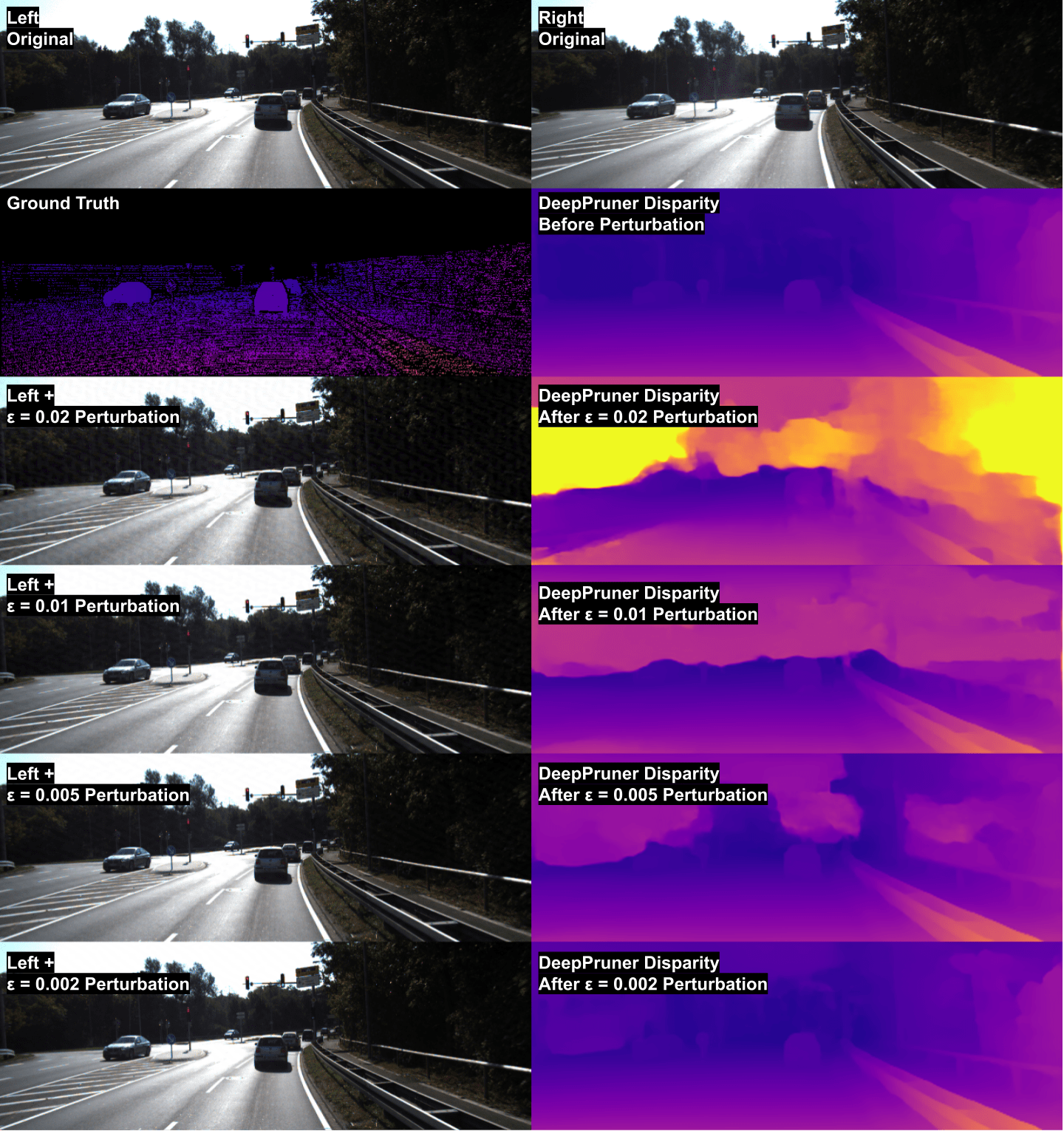}
\caption{\textit{Attacking DeepPruner at different upper norms $\epsilon \in \{0.002, 0.005, 0.01, 0.02\}$ for a scene from KITTI 2015}.}
\label{fig:deeppruner_kitti2015_different_norms}
\end{figure*}

\begin{figure*}[ht]
\centering
\includegraphics[width=1.0\textwidth]{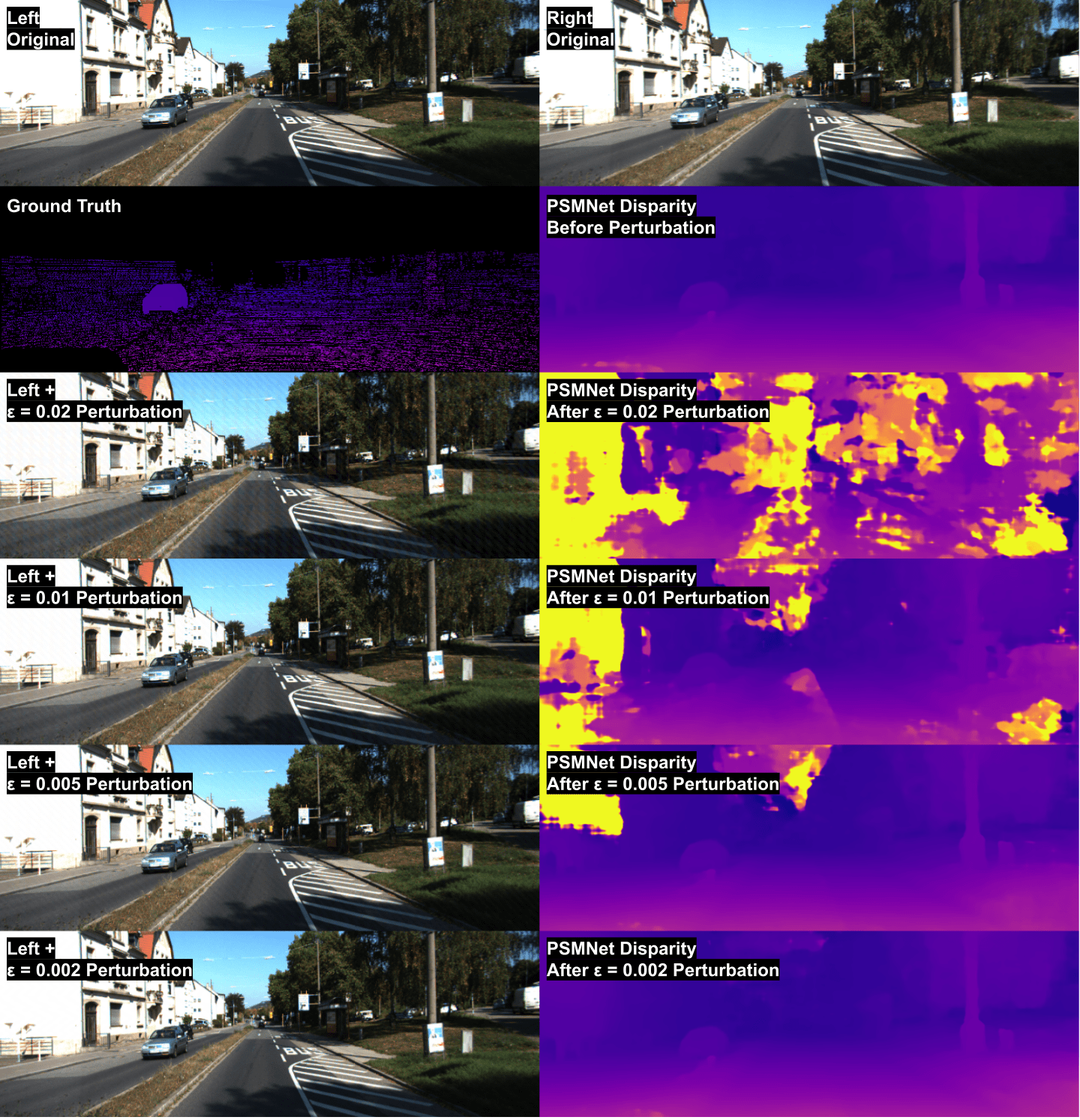}
\caption{\textit{Attacking PSMNet at different upper norms $\epsilon \in \{0.002, 0.005, 0.01, 0.02\}$ for a scene from KITTI 2015}.}
\label{fig:psmnet_kitti2015_different_norms}
\end{figure*}

\begin{figure*}[ht]
\centering
\includegraphics[width=1.0\textwidth]{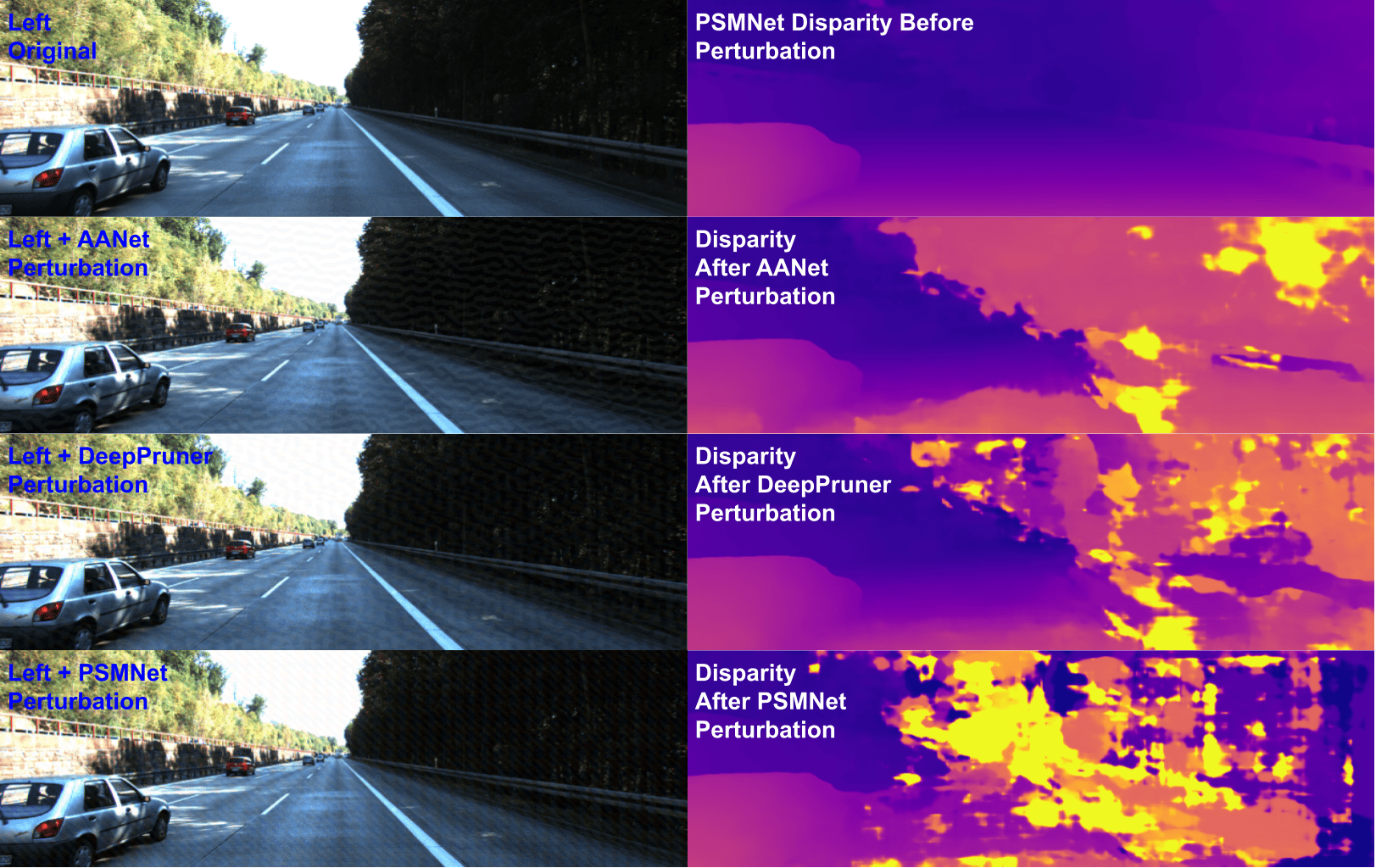}
\caption{\textit{Attacking PSMNet for a scene from KITTI 2015 with a stereoscopic universal perturbations optimized on KITTI for AANet, DeepPruner, and PSMNet}.}
\label{fig:kitti2015_to_psmnet}
\end{figure*}

\begin{figure*}[ht]
\centering
\includegraphics[width=1.0\textwidth]{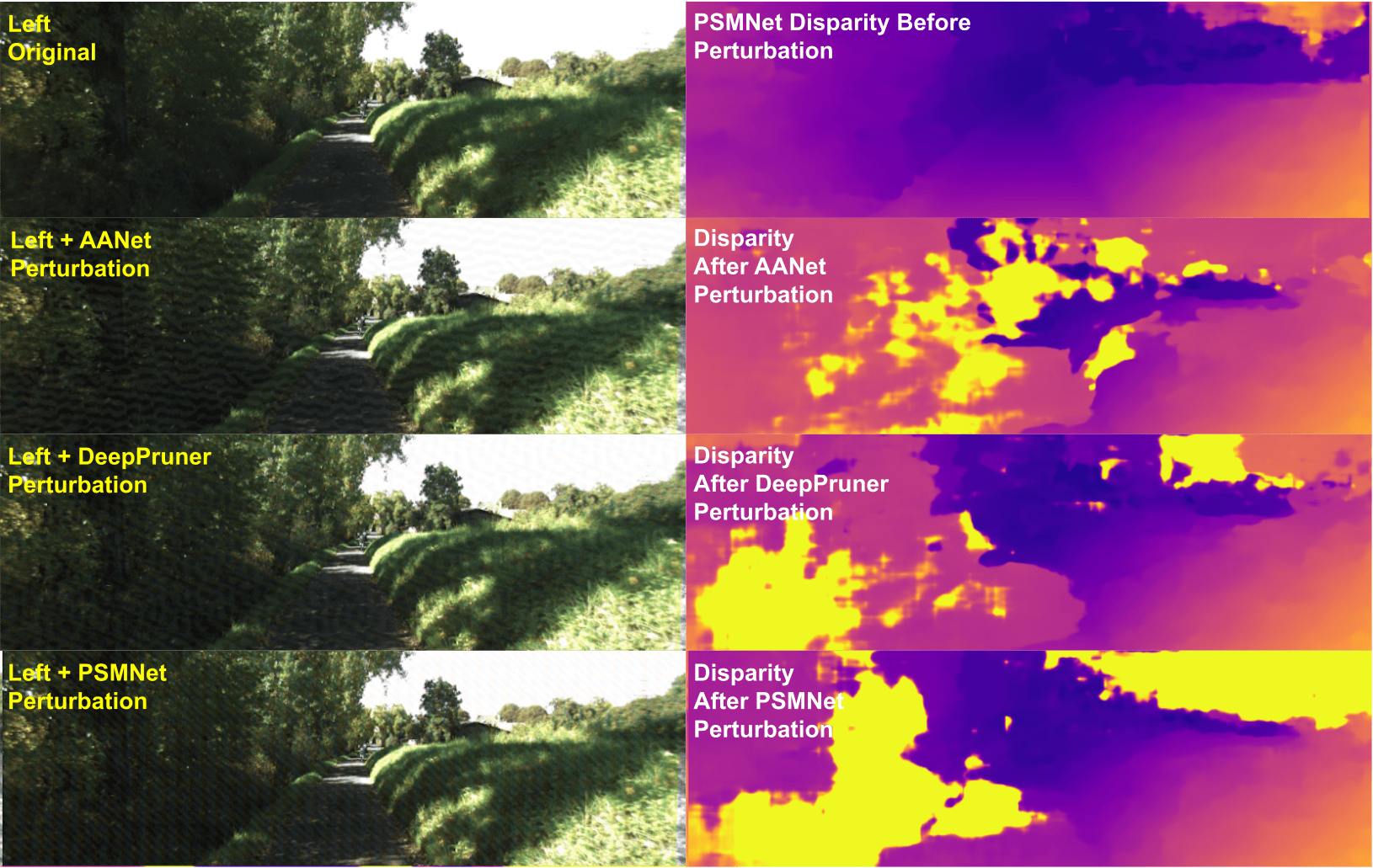}
\vspace{-1em}
\caption{\textit{Attacking PSMNet for a scene from KITTI 2012 with a stereoscopic universal perturbations optimized on KITTI for AANet, DeepPruner, and PSMNet}.}
\vspace{-1em}
\label{fig:kitti2012_to_psmnet}
\end{figure*}

\begin{figure*}[ht]
\centering
\includegraphics[width=1.0\textwidth]{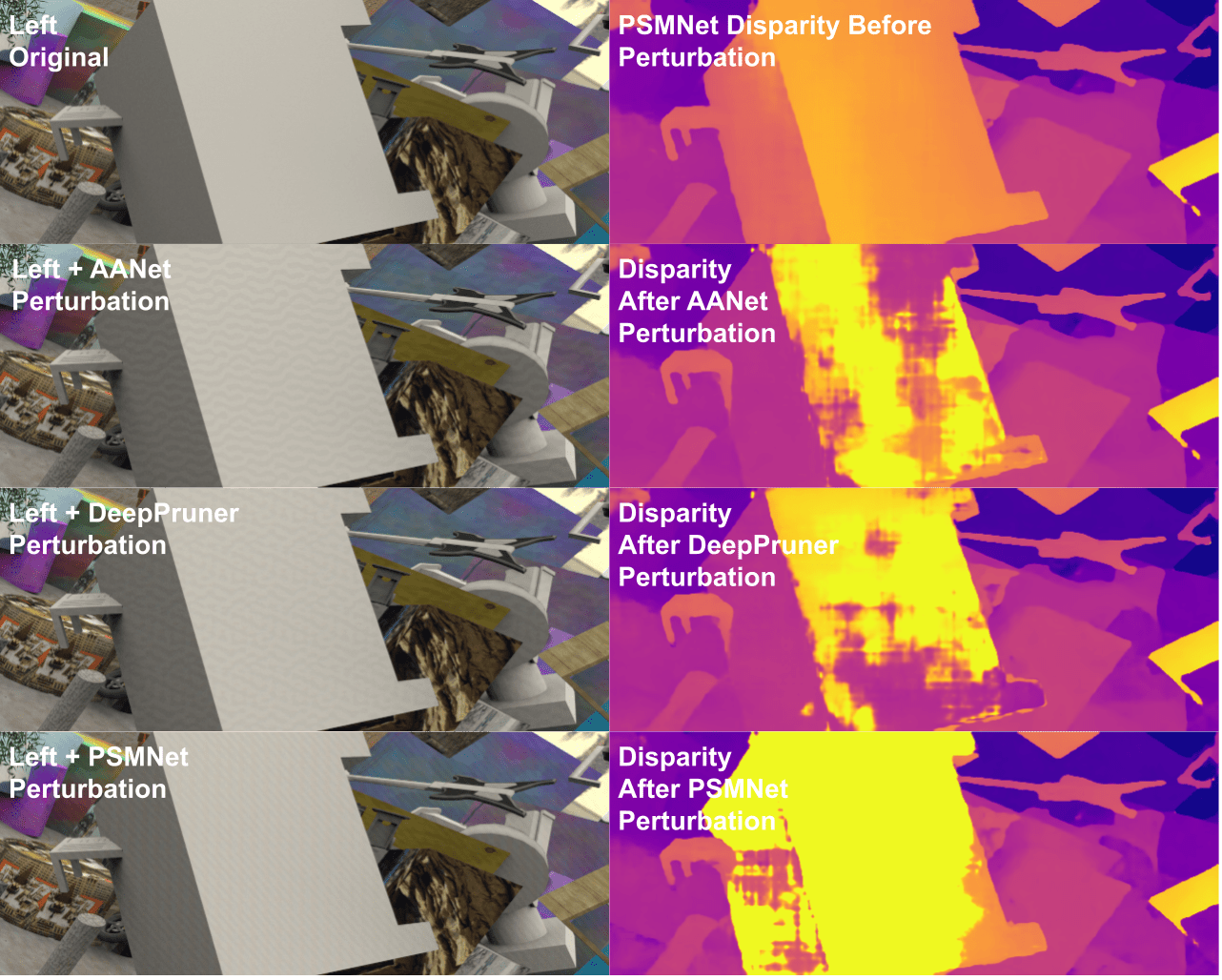}
\vspace{-1em}
\caption{\textit{Attacking PSMNet for a scene from FlyingThings3D with a stereoscopic universal perturbations optimized on KITTI for AANet, DeepPruner, and PSMNet}.}
\vspace{-1em}
\label{fig:flyingthing3d_to_psmnet}
\end{figure*}

Next, we demonstrate attacks against each model using the SUPs trained for it. We show the perturbed stereo pair, the original disparities estimated from clean stereo pairs and the corrupted disparities estimated from the perturbed stereo pair for KITTI 2015 (\figref{fig:kitti2015_attack}), KITTI 2012 (\figref{fig:kitti2012_attack}), and FlyingThings3D (\figref{fig:flyingthing3d_attack}). To demonstrate how an attack varies for different upper norms, we attack each network for scenes from KITTI 2015 using perturbations with upper norm $\epsilon \in \{0.002, 0.005, 0.01, 0.02\}$; we look at AANet in \figref{fig:aanet_kitti2015_different_norms}, DeepPruner in \figref{fig:deeppruner_kitti2015_different_norms}, and PSMNet in \figref{fig:psmnet_kitti2015_different_norms}. As expected, as the upper norm $\epsilon$ increases, we also observe more corruption in the disparity map. The corrupted regions are generally estimated as ``closer'' to the camera. We conclude by visualizing transferability of the SUPs to PSMNet across different datasets in \figref{fig:kitti2015_to_psmnet}, \figref{fig:kitti2012_to_psmnet}, and \figref{fig:flyingthing3d_to_psmnet}.

\end{appendices}

\end{document}